\documentclass{article}
\usepackage[utf8]{inputenc}
\usepackage[%
  style     = nature,
  autocite  = superscript,
  backend   = bibtex,
  sortcites = true,
]{biblatex}
\usepackage{amsmath}
\usepackage{booktabs}
\usepackage{graphicx}
\usepackage{caption}
\usepackage{subcaption}
\usepackage{longtable}
\usepackage[a4paper, margin=1in]{geometry}
\usepackage{caption}
\usepackage{float}
\usepackage{multirow, array}
\usepackage{authblk}
\usepackage{textcomp}
\usepackage{fancyhdr}
\usepackage{hyperref}

\graphicspath{{./figures/}}
\title{A Machine Learning Challenge for Prognostic Modelling in Head and Neck Cancer Using Multi-modal Data}

\author[1,2]{Michal Kazmierski}
\author[1,2,5]{Mattea Welch}
\author[1,2]{Sejin Kim}
\author[1,3,5]{Chris McIntosh}
\author[3]{Princess Margaret Head and Neck Cancer Group}
\author[3]{Katrina Rey-McIntyre}
\author[3,4]{Shao Hui Huang}
\author[3,5]{Tirth Patel}
\author[3,4]{Tony Tadic}
\author[3,4,5]{Michael Milosevic}
\author[3,4]{Fei-Fei Liu}
\author[3,4]{Andrew Hope}
\author[1,3,4]{Scott Bratman}
\author[1,2]{Benjamin Haibe-Kains\thanks{Corresponding author.}}

\affil[1]{Department of Medical Biophysics, University of Toronto, Toronto, ON}
\affil[2]{Princess Margaret Cancer Centre, Toronto, ON}
\affil[3]{Radiation Medicine Program, Princess Margaret Cancer Centre, Toronto, ON}
\affil[4]{Department of Radiation Oncology, University of Toronto, ON}
\affil[5]{TECHNA Institute, Toronto, ON}
\date{}

\fancypagestyle{preprintfoot}{%
  \fancyhf{}
  
  \fancyfoot[L]{Preprint. Under review.}
  \fancyfoot[C]{\thepage}
}

\begin{document}

\maketitle

\thispagestyle{preprintfoot}
\section*{Abstract}
Accurate prognosis for an individual patient is a key component of precision
oncology. Recent advances in machine learning have enabled the development of
 models using a wider range of data, including imaging. Radiomics
aims to extract quantitative predictive and prognostic biomarkers from routine
medical imaging, but evidence for computed tomography radiomics for
prognosis remains inconclusive. We have conducted an institutional machine learning challenge
to develop an accurate model for overall survival prediction in head and
neck cancer using clinical data etxracted from electronic medical records and pre-treatment
radiological images, as well as to evaluate the true added benefit of radiomics for head and neck cancer prognosis. Using a large, retrospective dataset of 2,552 patients and a rigorous
evaluation framework, we compared 12 different submissions using imaging and clinical
data, separately or in combination. The winning approach used non-linear,
multitask learning on clinical data and tumour volume, achieving high prognostic accuracy
for 2-year and lifetime survival prediction and outperforming models relying on
clinical data only, engineered radiomics and deep learning. Combining all submissions in
an ensemble model resulted in improved accuracy, with the highest gain from a
image-based deep learning model. Our results show the potential of machine learning and
simple, informative prognostic factors in combination with large datasets as a
tool to guide personalized cancer care.
\section*{Introduction}
The ability of computer algorithms to assist in clinical oncology tasks, such as
patient prognosis, has been an area of active research since the
1970's~\autocite{koss_computer-aided_1971}. More recently, machine learning (ML)
and artificial intelligence (AI) have emerged as a potential solution to process clinical data from
multiple sources and aid diagnosis~\autocite{ardila_end--end_2019},
prognosis~\autocite{chang_oral_2013} and course of
treatment decisions~\autocite{mahadevaiah_artificial_2020}, enabling a more precise
approach to clinical management taking individual patient characteristics into
account~\autocite{shrager_rapid_2014}. The need for more personalized care is
particularly evident in head and neck cancer (HNC), which exhibits significant
heterogeneity in clinical presentation, tumour biology and
outcomes~\autocite{pai_molecular_2009,leemans_molecular_2011}, making it difficult to select the optimal management strategy for each patient. Hence, there is a current need for
better predictive and prognostic tools to guide clinical decision
making~\autocite{mirghani_treatment_2015,osullivan_deintensification_2013}.

One potential source of novel prognostic information is the
imaging data collected as part of standard care. Imaging data has the potential to increase the scope of relevant prognostic factors in a non-invasive manner as compared to genomics or pathology, while high
volume and intrinsic complexity render it an excellent use case for machine
learning. \textit{Radiomics} is an umbrella term for the emerging field of
research aiming to develop new non-invasive quantitative prognostic and
predictive imaging biomarkers using both
hand-engineered~\autocite{gillies_radiomics:_2016} and deep learning
techniques~\autocite{lecun_deep_2015}. Retrospective radiomics studies have been
performed in a variety of imaging modalities and cancer types for a range of
endpoints.

In HNC, radiomics has been
used~\autocite{wong_radiomics_2016} to predict patient
outcomes~\autocite{ger_radiomics_2019,lv_multi-level_2020}, treatment
response~\autocite{vallieres_radiomics_2017,diamant_deep_2019},
toxicity~\autocite{sheikh_predicting_2019,van_dijk_delta-radiomics_2019}, and
discover associations between imaging and genomic
markers~\autocite{huang_development_2019,zhu_imaging-genomic_2019}. In a recent MICCAI Grand Challenge, teams from several different institutions
competed to develop the best ML approach to predict human papillomavirus (HPV)
status using a public HNC
dataset~\autocite{miccaimd_anderson_cancer_center_head_and_neck_quantitative_imaging_working_group_matched_2017}.

Despite the large number of promising retrospective studies, the adoption of
prognostic models utilizing radiomics into clinical workflows is
limited~\autocite{sanduleanu_tracking_2018,morin_deep_2018}. There has also been
growing concern regarding the lack of transparency and reproducibility in ML
research, which are crucial to enabling widespread
adoption of these tools~\autocite{hutson_artificial_2018,
  bluemke_assessing_2020}. Although significant progress has been made in
certain areas (e.g. ensuring consistency between different engineered feature
toolkits~\autocite{zwanenburg_image_2020}) many studies do not make the code or
data used for model development publicly available, and do not report key
details of data processing, model training and validation, making it challenging
to reproduce, validate and build upon their
findings~\autocite{sanduleanu_tracking_2018, haibe-kains_transparency_2020}.
Additionally, the lack of large, standardized benchmark datasets makes comparing
different approaches challenging, with many publications relying on small,
private datasets. Furthermore, there is mounting evidence that the current
methods of image quantification based on engineered features are largely
correlated and redundant to accepted clinical
biomarkers~\autocite{welch_vulnerabilities_2019, ger_radiomics_2019,
  traverso_machine_2020}. Approaches based on deep learning have been steadily
gaining popularity, but it is still unclear whether they share these pitfalls and
truly improve prognostic performance.

We have conducted an institutional ML challenge to develop a
prognostic model for HNC using routine CT imaging and data from electronic
medical records (EMR), engaging a group of participants from diverse academic
background and knowledge. In this manuscript we will describe in detail the
challenge framework, prognostic model submissions and final results. A key
strength of our approach is the rigorous evaluation framework, enabling rigorous
comparison of multiple ML approaches using EMR data as well as engineered and
deep radiomics in a large dataset of 2,552 HNC patients. We make the dataset and
the code used to develop and evaluate the challenge submissions publicly
available for the benefit of the broader community.

\section*{Results}
\subsection*{Dataset}
\begin{figure}[H]
  \centering
  \begin{subfigure}[t]{\textwidth}
    % \caption{}
    \centering \includegraphics[width=\textwidth]{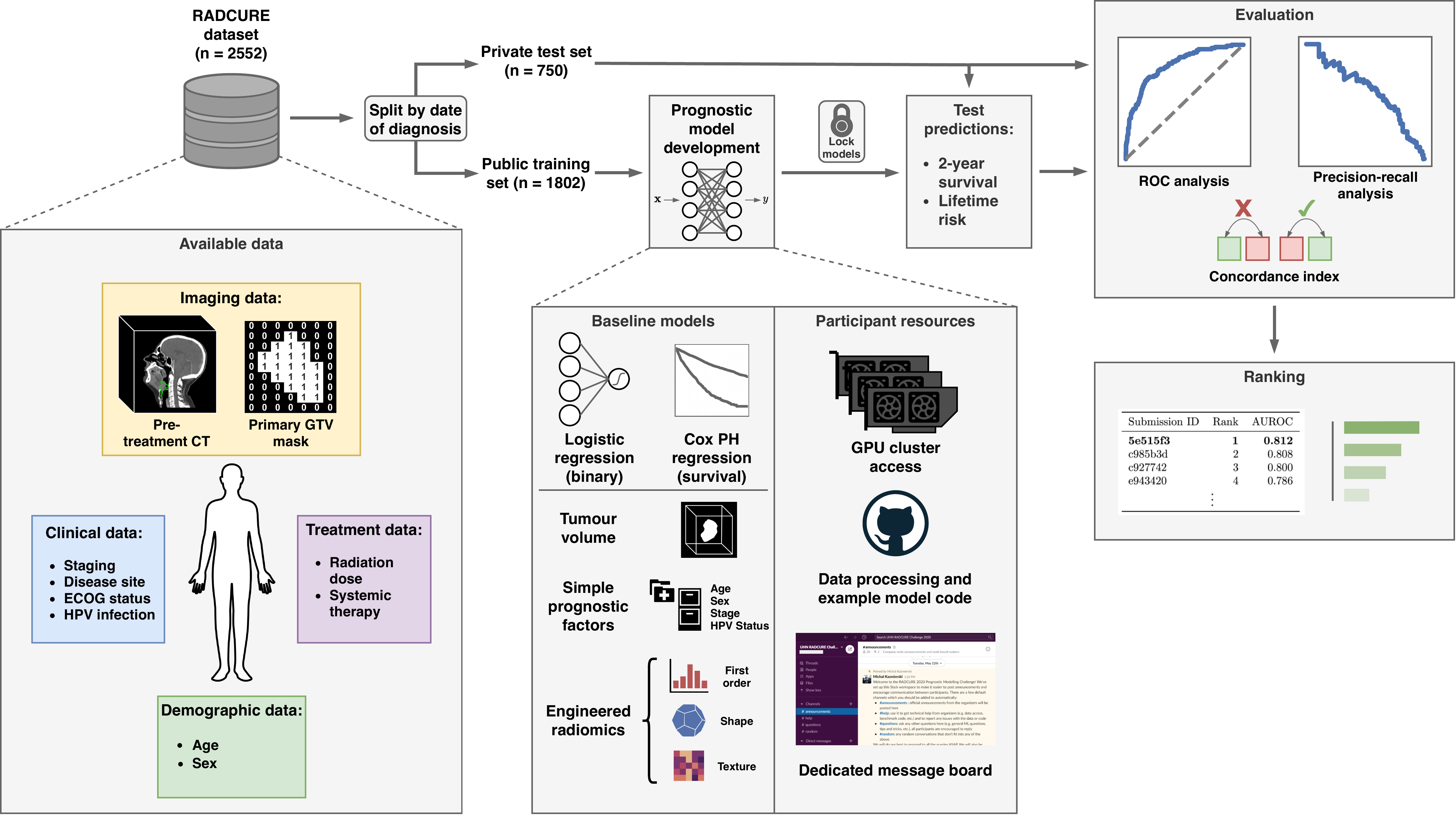}
  \end{subfigure}
  \caption{\textbf{Overview of challenge and dataset.} EMR and imaging data from a large cohort of HNC patients were
    available to the participants. The training set, consisting of patients
    diagnosed before a pre-specified date was released together with the
    ground-truth outcome information and used for prognostic model development.
    The test set was kept private and only made available (without outcome data)
    after the development phase was completed. The participants submitted their test
    set predictions, which were evaluated by the organizers. We provided the
    participants with a rich set of resources, including access to a computing
    cluster with general purpose graphics processing units (GPUs), a Github
    repository containing the code used for data processing and submission
    evaluation together with example model implementation, as well as a
    dedicated Slack workspace used for announcements, communication between
    participants and organizers and technical support. We also developed a set
    of simple but strong baseline models to serve as benchmark for comparison
    as well as a reference point for the participants during the development
    stage.}

  \label{fig:challenge_overview}
\end{figure}

For the purpose of the challenge, we collected a combined EMR (i.e. clinical, demographic and interventional data) and imaging
dataset of 2,552 HNC patients treated with radiation therapy or chemoradiation at
Princess Margaret Cancer Centre (PM). The dataset was
divided into a training set (70\%) and an independent test set (30\%). In order to
simulate a prospective validation we chose to split the data by date of
diagnosis at a pre-defined time point. We made pre-treatment contrast-enhanced CT
images and binary masks of primary gross tumour volumes (GTV) available to the
participants. We also released a set of variables extracted from EMR, including
demographic (age at diagnosis, sex), clinical (T, N and overall stage, disease
site, performance status and HPV infection status) and treatment-related
(radiation dose in Gy, use of chemotherapy) characteristics
(fig.~\ref{fig:challenge_overview}). Additionally, outcome data (time to death
or censoring, event indicator) was available for training data only.

\subsection*{Challenge description and evaluation criteria}
The challenge organization process is summarized in fig.~\ref{fig:challenge_overview}. The challenge was open to anyone within the University Health Network. Study
design and competition rules were fully specified and agreed upon by all
participants prior to the start. All participants had access to the
training data with ground-truth outcome labels, while the test set was held out
for final evaluation. The primary objective was to predict 2-year overall
survival (OS), with the secondary goals of predicting a patient's lifetime risk
of death and full survival curve. We chose the binary endpoint as it is commonly
used in the literature and readily amenable to many standard ML methods. The
primary evaluation metric for the binary endpoint, which was used to rank the
submissions, was the area under receiver operating characteristic curve (AUROC).
We also used average precision (AP) as a secondary performance measure to break
any submission ties, due to its higher sensitivity to class
imbalance~\autocite{saito_precision-recall_2015}. Submission of lifetime risk
and survival curve predictions was optional, and they were scored using the
concordance (\(C\)) index~\autocite{harrell_multivariable_1996}. Importantly,
the participants were blinded to test set outcomes and only submitted
predictions to be evaluated by the organizers. We additionally created a set of
benchmark models for comparison (see Methods). We did not enforce any particular
model type, image processing or input data (provided it was part of the official
training set), although we did encourage participants to submit predictions
based on EMR features, images, and combined data separately (if they chose to
use all of the data modalities).

\subsection*{Overview of submissions}
\begin{figure}[H]
  \centering
  \begin{subfigure}[t]{.7\textwidth}
    \caption{}
    \centering \includegraphics[width=\textwidth]{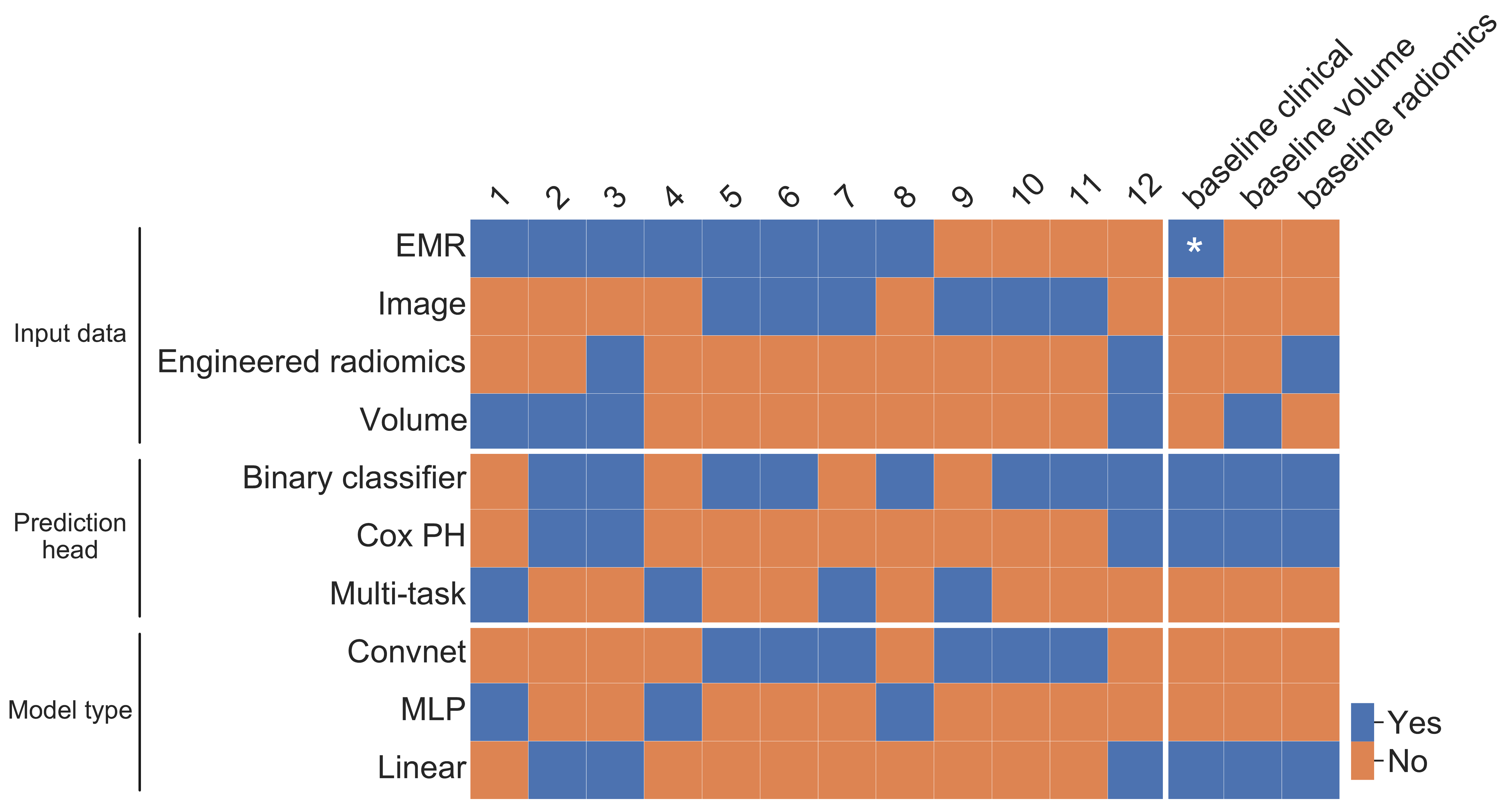}
    \label{fig:overview_heatmap}
  \end{subfigure}
  \vfill
  \begin{subfigure}[t]{.323\textwidth}
    \caption{}
    \centering \includegraphics[width=\textwidth]{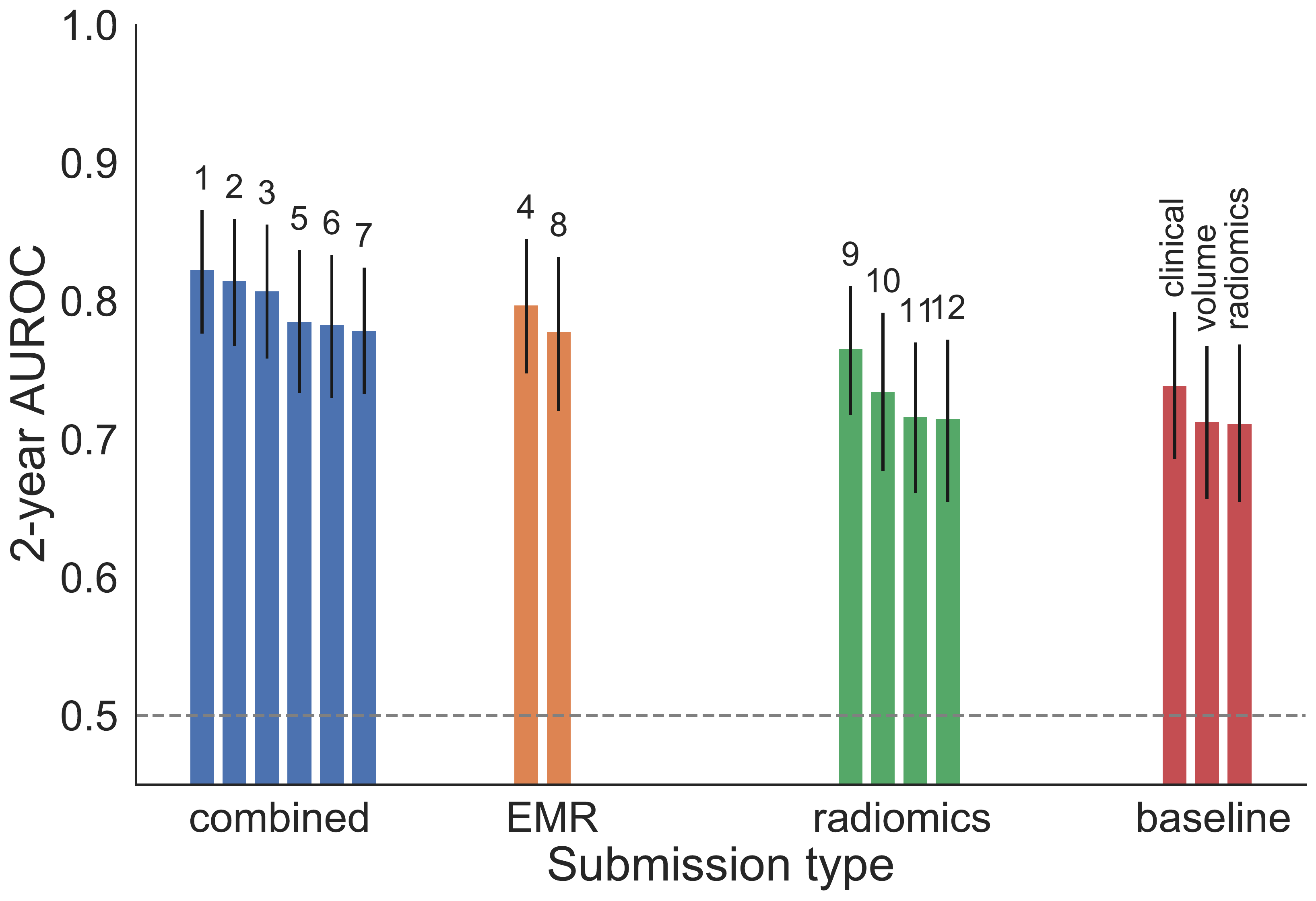}
  \end{subfigure}
  \hfill
  \begin{subfigure}[t]{.323\textwidth}
    \caption{}
    \centering \includegraphics[width=\textwidth]{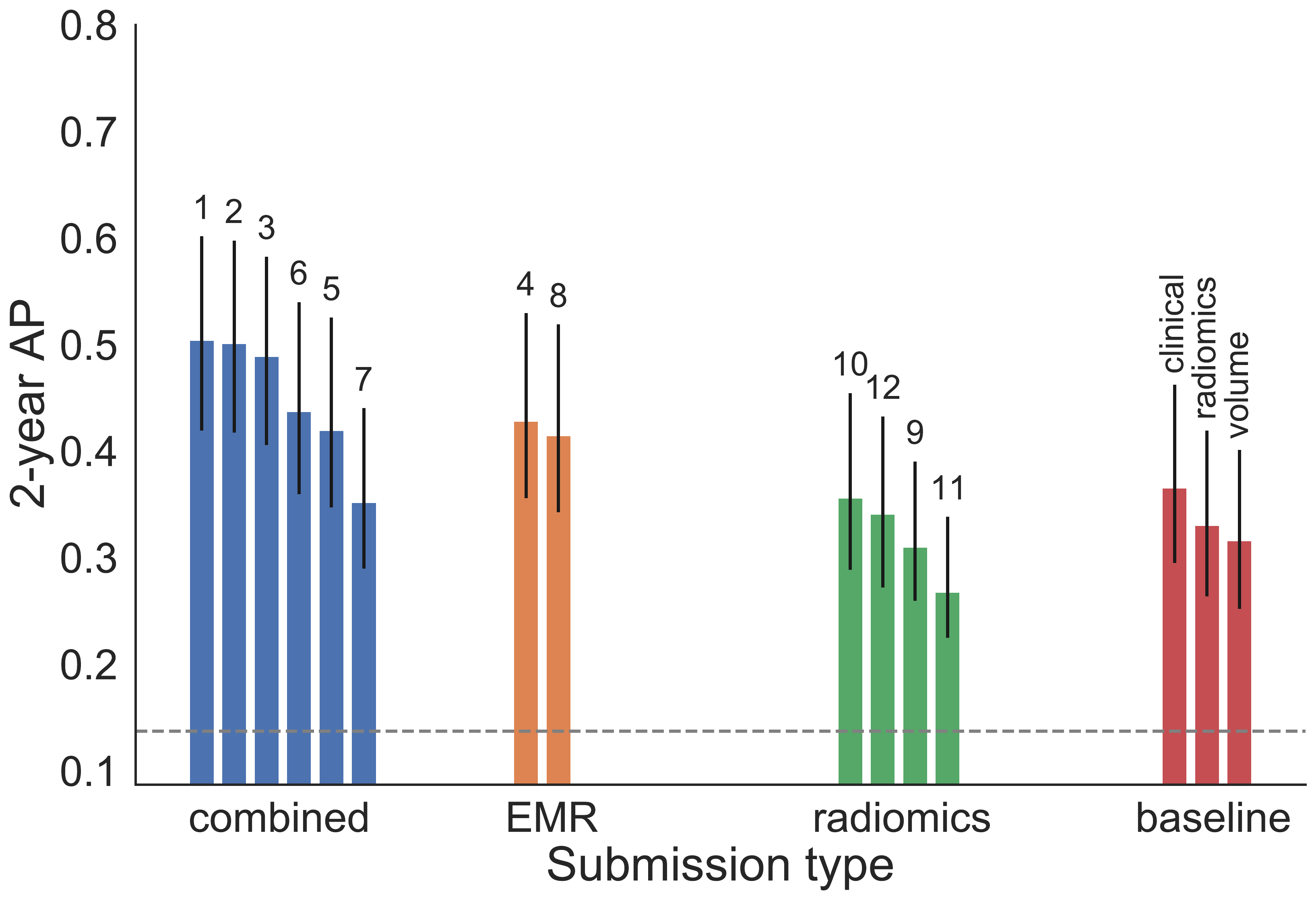}
  \end{subfigure}
  \hfill
  \begin{subfigure}[t]{.323\textwidth}
    \caption{}
    \centering \includegraphics[width=\textwidth]{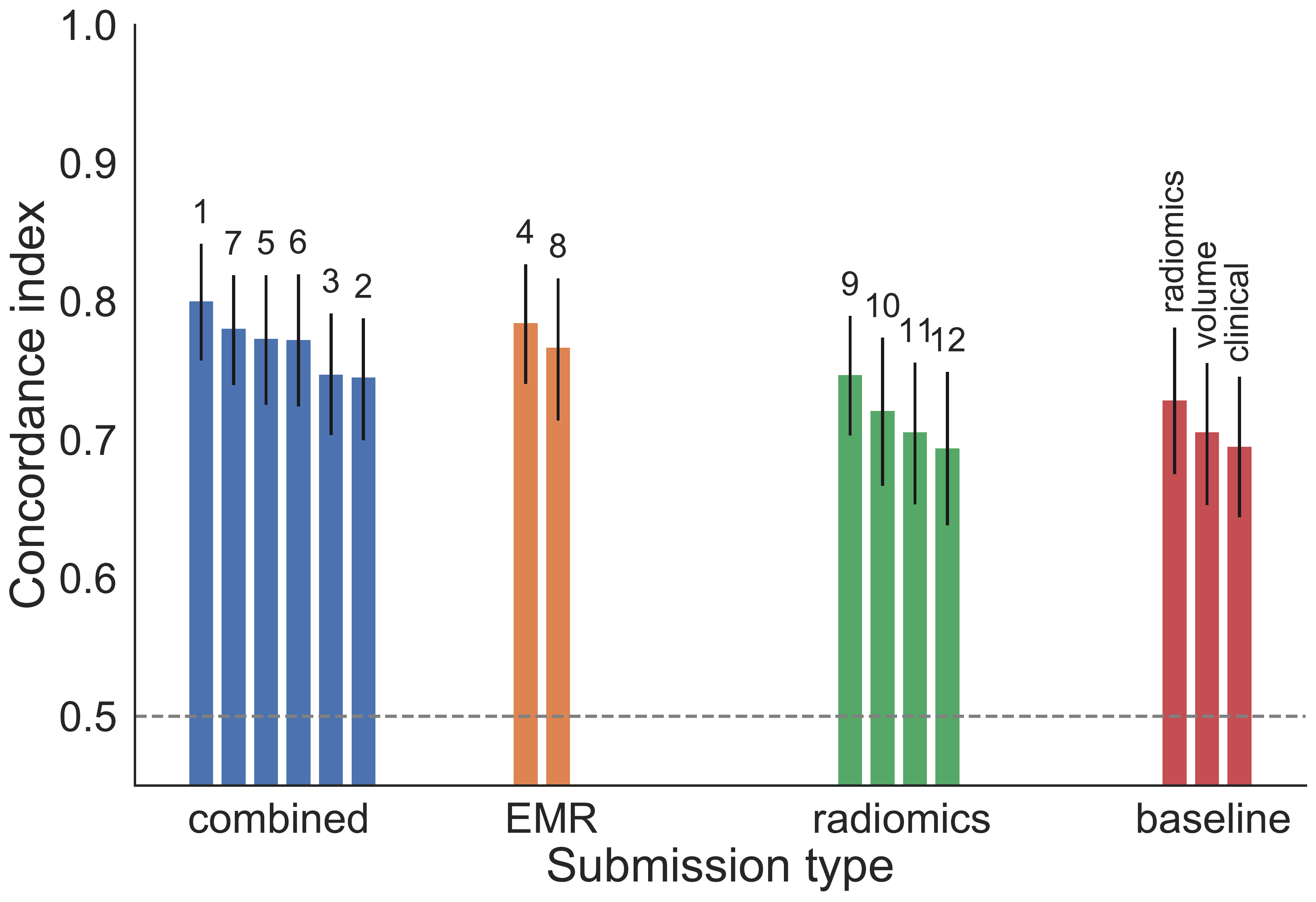}
  \end{subfigure}
    \vfill
  \begin{subfigure}[t]{.323\textwidth}
    \caption{}
    \centering \includegraphics[width=\textwidth]{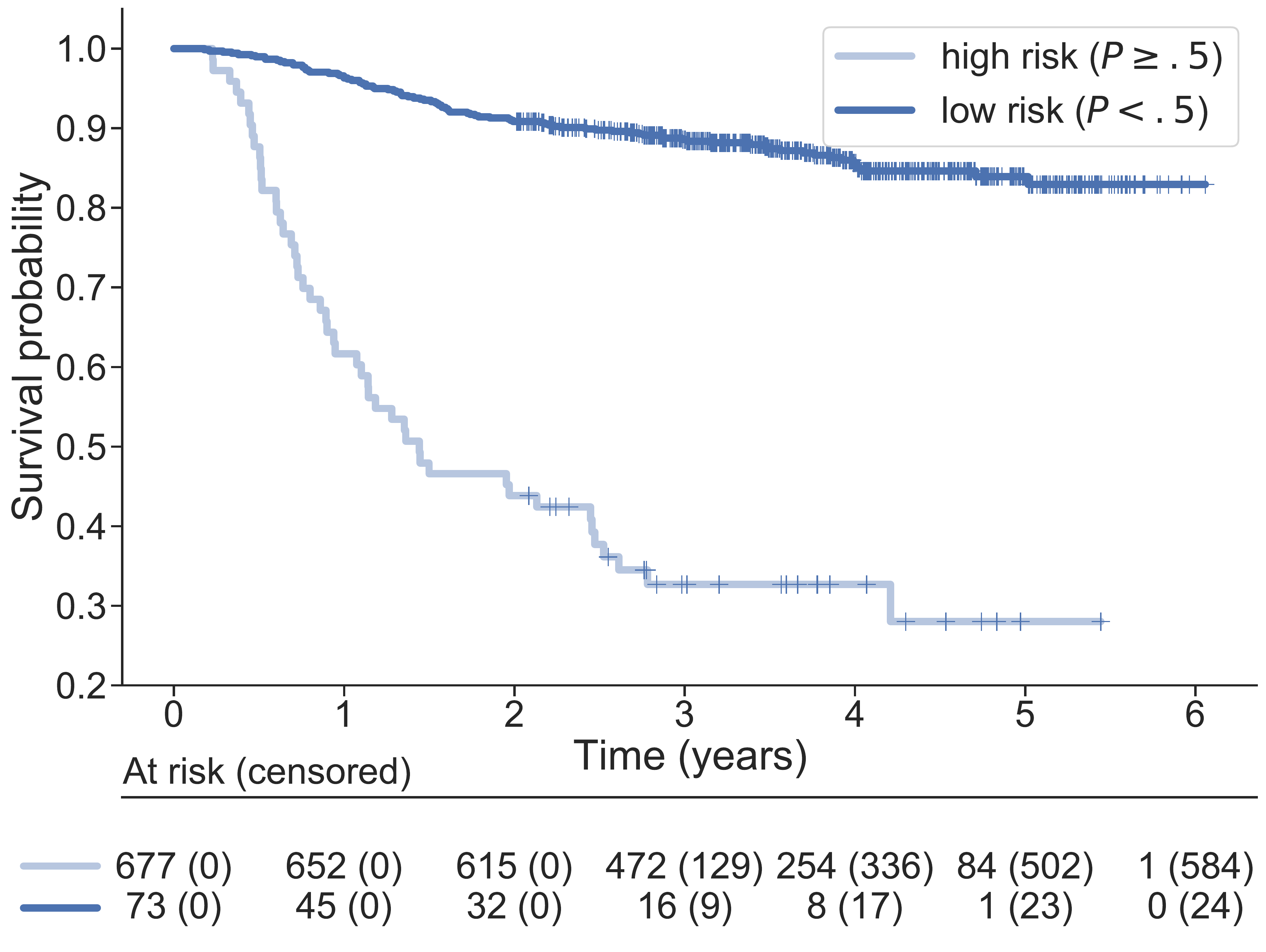}
  \end{subfigure}
  \begin{subfigure}[t]{.323\textwidth}
    \caption{}
    \centering \includegraphics[width=\textwidth]{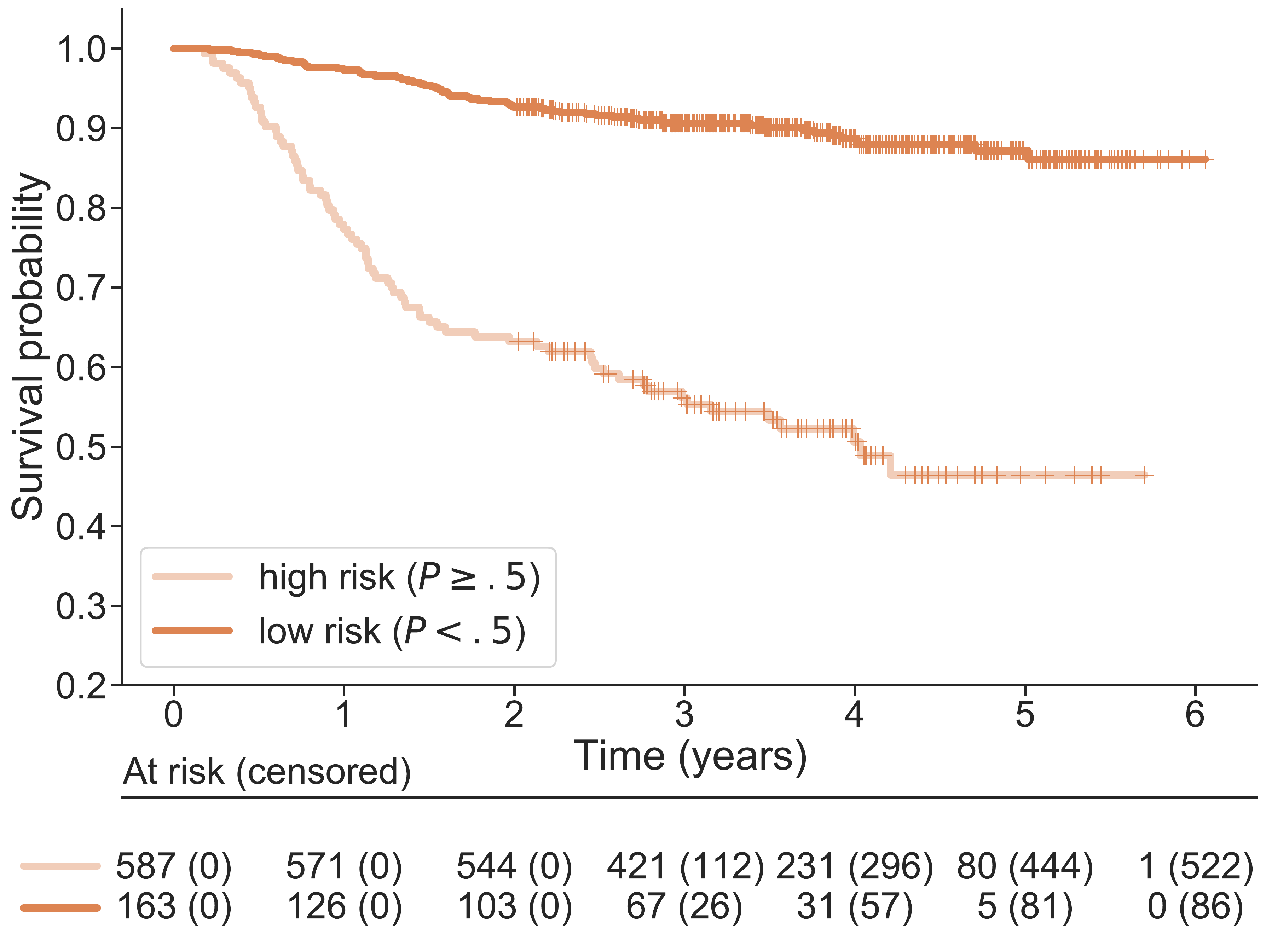}
  \end{subfigure}
  \begin{subfigure}[t]{.323\textwidth}
    \caption{}
    \centering \includegraphics[width=\textwidth]{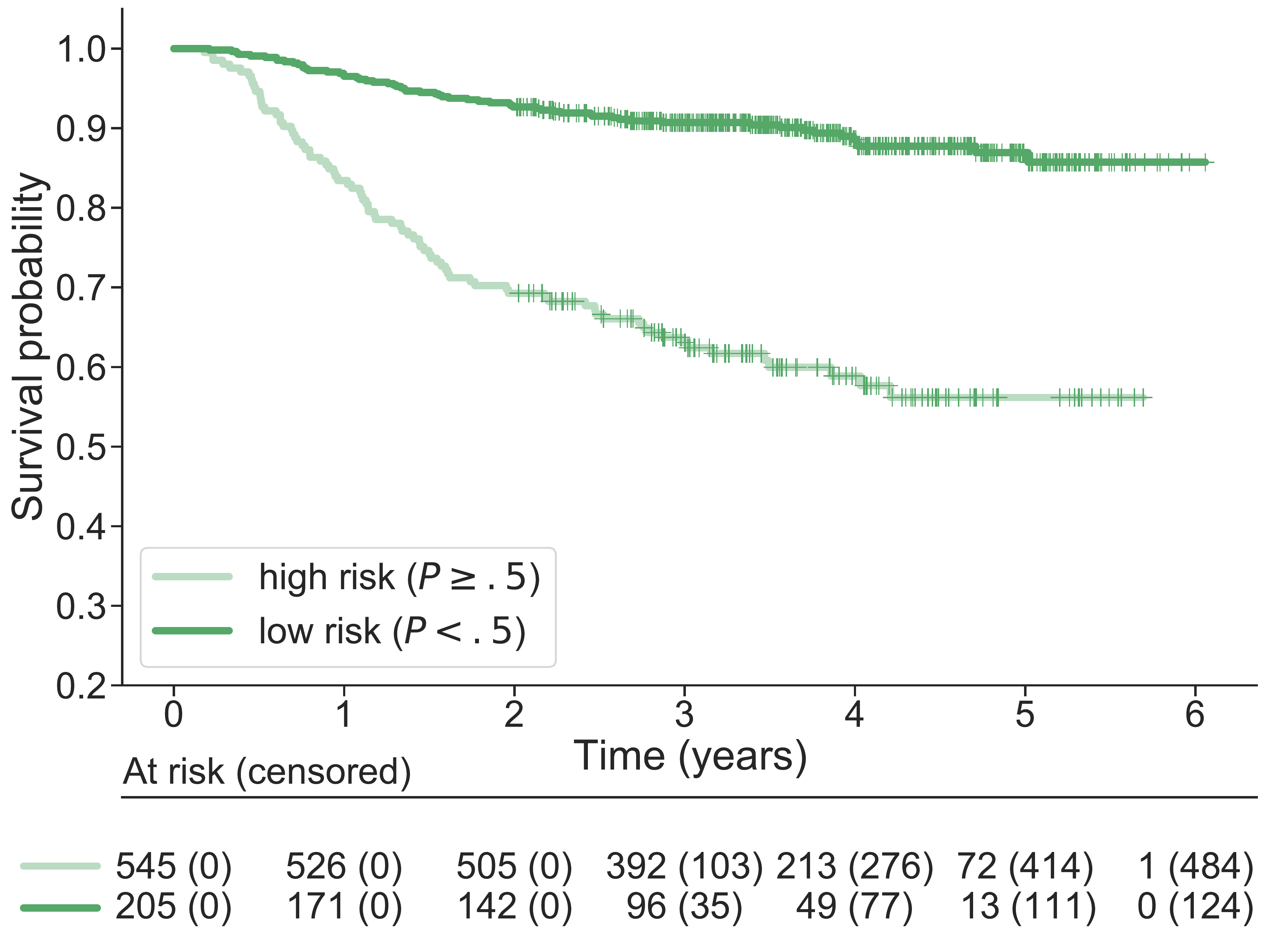}
  \end{subfigure}
  \caption{\textbf{Challenge results.}  \textbf{(a)} Overview of submission
    characteristics. The characteristics are grouped into input data, prediction
    head (i.e. how were the survival predictions made) and model type (whether
    the model involved any nonlinearities and/or convolutions). PH: proportional
    hazards, MLP: multilayer perceptron, *: age, sex, stage, HPV status.
    \textbf{(b,c,d)} Performance of all challenge submissions
    and benchmark models in terms of 2-year AUROC,
    2-year average precision and concordance index of the lifetime
    risk respectively. The results are ranked by AUROC (numbers above bars indicate the
    overall rank of each submission). Error bars represent 95\% confidence
    intervals computed using 10,000 stratified bootstrap replicates. Dashed grey
    lines indicate random guessing performance (.5 for AUROC and \(C\)-index,
    .14 for AP). \textbf{(e,f,g)} show the Kaplan-Meier survival estimates in
    low and high risk groups identified by the best performing model in each
    challenge category (combined, EMR only and radiomics) respectively. Test set
    patient were stratified into 2 groups based on the predicted 2-year event
    probability at .5 threshold. In each case, there were significant
    differences in survival between the predicted risk groups (hazard ratio
    8.64, 5.96 and 4.50 respectively, \(p < 10^{-18}\) for all).}
  \label{fig:metrics}
\end{figure}

We received 12 submissions in total, which can be broadly classified as using
EMR factors only, imaging only, or combining all data sources. In addition to the
required 2-year event probabilities, 10 submissions included lifetime risk
predictions and 7 included the full predicted survival curves. All submitted
models performed significantly better than random on all performance measures
(\(p < .0001\) by permutation test). The top submission performed significantly better in terms of  AUROC than every other submission (\(\mathrm{FDR} < .05\)), except the second-best (\(\mathrm{FDR} > .05\) Most participants who used the imaging data
relied on convolutional neural networks (convnets) to automatically learn
predictive representations; only 2 combined and 1 radiomics-only submission used
handcrafted features. Of the submitted convnets, 2 out of 3 relied on
three-dimensional (3D) convolution operations. Although all EMR-only approaches
used the same input data, there was significant variation in the kind of model
used (linear and nonlinear, binary classifiers, proportional hazards and
multi-task models, fig. \ref{fig:overview_heatmap}). The combined submissions used EMR data together with either
tumour volume (\(n = 2\)), engineered radiomics (\(n = 1\)) or deep learning (\(n = 3\)). A brief overview of all
submissions is presented in table~\ref{tab:results_summary}; for detailed
descriptions, see Supplementary Material.

% \begin{table}
%   \centering
\begin{scriptsize}
  \begin{longtable}{rp{.5\textwidth} p{.15\textwidth} p{.15\textwidth} p{.15\textwidth}}
  \toprule
  Rank &                                                                                                                                                                                                                Description &                 AUROC &                    AP &               \(C\)-index \\
  \midrule
    1 &                                                                                      Deep multi-task logistic regression using EMR features and tumour volume. &  0.823 [0.777--0.866] &  0.505 [0.420--0.602] &  0.801 [0.757--0.842] \\
    2 &                                 Fuzzy logistic regression (binary) and Cox proportional hazards model (risk prediction) using EMR features and tumour volume.  &  0.816 [0.767--0.860] &  0.502 [0.418--0.598] &  0.746 [0.700--0.788] \\
    3 &                    Fuzzy logistic regression (binary) or Cox proportional hazards model (risk prediction) using EMR features and engineered radiomic features. &  0.808 [0.758--0.856] &  0.490 [0.406--0.583] &  0.748 [0.703--0.792] \\
    4 &                                                                                                             Multi-task logistic regression using EMR features. &  0.798 [0.748--0.845] &  0.429 [0.356--0.530] &  0.785 [0.740--0.827] \\
    5 &                                      3D convnet using cropped image patch around the tumour with EMR features concatenated before binary classification layer. &  0.786 [0.734--0.837] &  0.420 [0.347--0.525] &  0.774 [0.725--0.819] \\
    6 &  2D convnet using largest GTV image and contour slices with EMR features concatenated after additional non-linear encoding before binary classification layer. &  0.783 [0.730--0.834] &  0.438 [0.360--0.540] &  0.773 [0.724--0.820] \\
    7 &                                     3D DenseNet using cropped image patch around the tumour with EMR features concatenated before multi-task prediction layer. &  0.780 [0.733--0.824] &  0.353 [0.290--0.440] &  0.781 [0.740--0.819] \\
    8 &                                                                  Multi-layer perceptron (MLP) with SELU activation and binary output layer using EMR features. &  0.779 [0.721--0.832] &  0.415 [0.343--0.519] &  0.768 [0.714--0.817] \\
    9 &                        Two-stream 3D DenseNet with multi-task prediction layer using cropped patch around the tumour and additional downsampled context patch. &  0.766 [0.718--0.811] &  0.311 [0.260--0.391] &  0.748 [0.703--0.790] \\
   10 &                                                                                 2D convnet using largest GTV image and contour slices and binary output layer. &  0.735 [0.677--0.792] &  0.357 [0.289--0.455] &  0.722 [0.667--0.774] \\
   11 &                                                                                3D convnet using cropped image patch around the tumour and binary output layer. &  0.717 [0.661--0.770] &  0.268 [0.225--0.339] &  0.706 [0.653--0.756] \\
   12 &                                    Fuzzy logistic regression (binary) and Cox proportional hazards model (risk prediction) using engineered radiomic features. &  0.716 [0.655--0.772] &  0.341 [0.272--0.433] &  0.695 [0.638--0.749] \\
  \bottomrule
  \caption{Summary of challenge submissions and performance metrics.}
  \label{tab:results_summary}
\end{longtable}
% \end{table}
\end{scriptsize}

\subsection*{Deep learning using imaging only achieves good performance and outperforms engineered radiomics}
Among the radiomics-only models, deep learning-based approaches
performed better than hand-engineered features. In particular, nearly all deep
learning models (except one) outperformed \texttt{baseline-radiomics} and
challenge submissions (submission 12, fig.~\ref{fig:metrics}) in the binary
prediction task (the smaller differences in \(C\)-index can be explained by the
fact that most of the deep models were designed for binary classification only,
and we used their binary predictions as a proxy for lifetime risk scores). We
note that it is difficult to draw definitive conclusions due to the large number
of radiomics toolkits and the wealth of feature types and configuration options
they offer~\autocite{zwanenburg_image_2020}. Nevertheless, the results show that a
carefully-tuned DL model can learn features with superior discriminative power
given a sufficiently large dataset.

The convnet-based models show varying levels of performance, most likely due to
differences in architectures and prior image processing. Notably, the best 3D
architecture (submission 9) achieves superior performance to the 2D VGGNet
(number 10). It incorporates several innovative features, including dense
connectivity~\autocite{huang_densely_2016, de_fauw_clinically_2018}, two-stream
architecture with a downsampled context window around the tumour and a dedicated
survival prediction head (detailed description in Supplementary Material).

\subsection*{EMR features show better prognostic value than deep learning, even in combination}
While radiomics can be a strong predictor of survival on its own, the small
performance gap between EMR and combined models using deep radiomics
(submissions 4, 6, 7) in most cases suggests the models do not learn
complementary image representations and that the performance is driven primarily
by the EMR features (fig.~\ref{fig:metrics}). Although one combined submission
using engineered features (number 3) achieved good performance, it performed
worse than the exact same model using EMR features and volume only (number 2),
indicating that adding radiomic features reduces performance, and the engineered
features were not strong predictors on their own. Moreover, none of the
radiomics-only models performed better than any of the EMR-only submissions
(although one convnet did outperform \texttt{baseline-clinical}). A possible
explanation is suboptimal model design that fails to exploit the complementarity
between the data sources. All of the deep learning solutions incorporated EMR
features in a rather ad hoc fashion by concatenating them with the image
representation vector and passing them to the final classification layer. While
this approach is widely used, it is not clear that it is optimal in this
context. More sophisticated methods of incorporating additional patient-level
information, such as e.g. joint latent spaces~\autocite{cheerla_deep_2019}
should be explored in future research.

\subsection*{Impact of volume dependence on model performance}
\begin{figure}[H]
  \centering
  \begin{subfigure}[t]{.48\textwidth}
    \caption{}
    \centering \includegraphics[width=\textwidth]{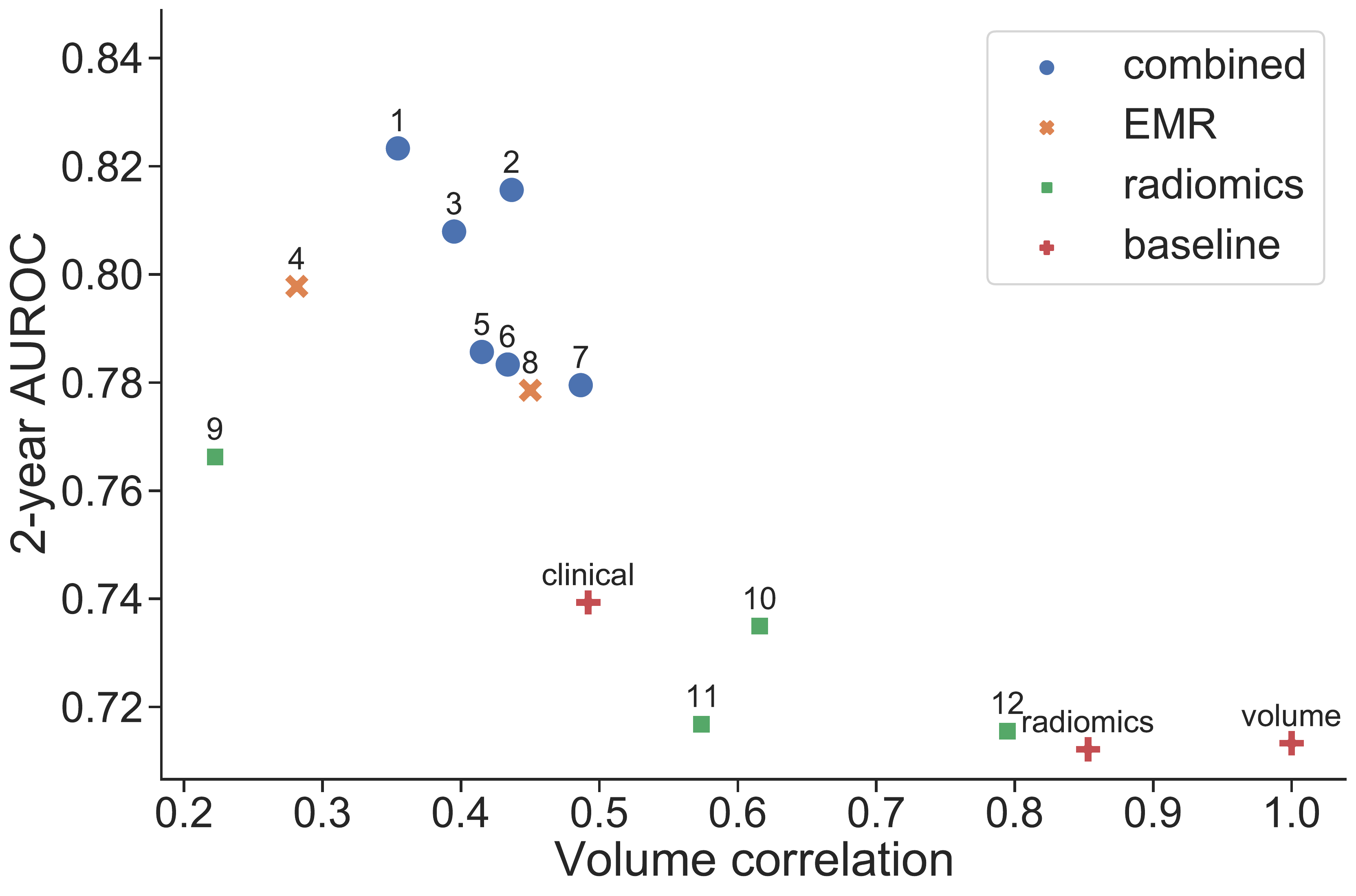}
  \end{subfigure}
  \hfill
  \begin{subfigure}[t]{.48\textwidth}
    \caption{}
    \centering \includegraphics[width=\textwidth]{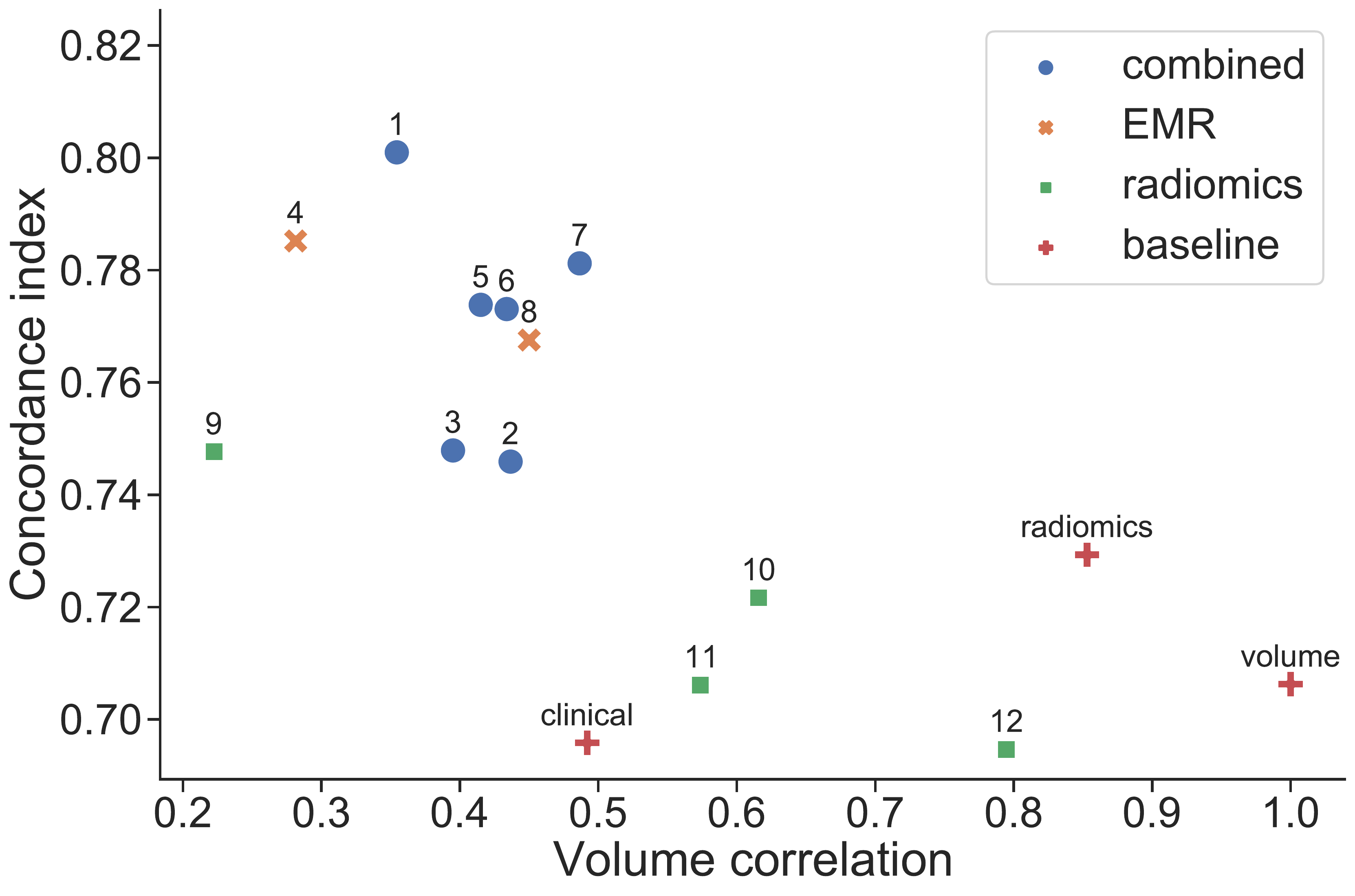}
  \end{subfigure}
  \caption{\textbf{Volume dependence of predictions.} Spearman rank correlation
    of the predictions of each submission with tumour volume against performance
    in terms of \textbf{(a)} AUROC and \textbf{(b)} \(C\)-index, respectively.
    The top submissions fall into an optimal region of low (but non-zero) volume
    correlation and high performance. Note that while submissions 1 and 2 used
    tumour volume as one of the input variables, their predictions correlate
    with volume only moderately (\(\rho < .5\)), indicating they are able to exploit additional
    information present in the EMR features. Higher correlation leads to
    decreased performance as the predictions are increasingly driven by volume
    only. Most radiomics-only submission fall in the high correlation region
    (\(\rho \ge .5\)), although deep learning predictions correlate at notably lower
    level than engineered features. Interestingly, the best radiomics submission
    (number 9) achieves the lowest volume correlation, suggesting that it might
    be using volume-independent imaging characteristics.}
  \label{fig:volume_corr}
\end{figure}

Recent literature has demonstrated that many radiomic signatures show strong
dependence on tumour volume~\autocite{welch_vulnerabilities_2019,
  traverso_machine_2020}, which is a simple image-derived feature and a known
prognostic factor in HNC~\autocite{lin_tumor_2017}. We evaluated the correlation
of all binary predictions with volume using Spearman rank correlation
(fig.~\ref{fig:volume_corr}). Both the baseline radiomics model and the
submission using handcrafted features show high correlation (Spearman
\(\rho=.79\) and \(\rho=.85\), respectively), suggesting that their predictions
are driven primarily by volume dependency. The predictions of two out of three
convnets also show some volume correlation, albeit smaller than engineered
features (\(\rho > .5\)). Interestingly, predictions of the best radiomics-only
model (submission 9) show only weak correlation with volume (\(\rho=.22\)) and
are more discriminative than volume alone (\(\mathrm{AUROC} = .77\)), suggesting
that it might be possible to learn volume-independent image-based predictors.

\subsection*{Winning Submission: Multi-task learning with simple image features and EMR data}
\begin{figure}[H]
  \centering
  \begin{subfigure}[t]{.75\textwidth}
    \caption{}
    \centering \includegraphics[width=\textwidth]{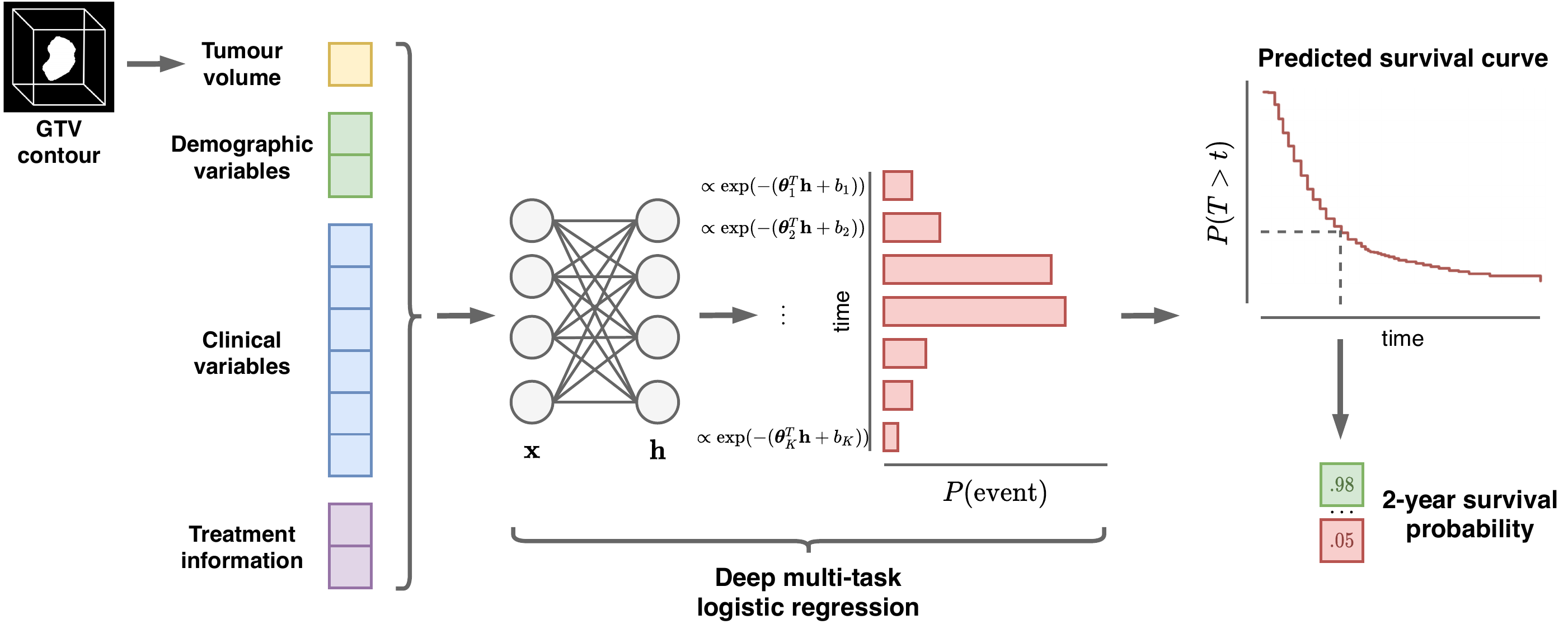}
  \end{subfigure}
  \begin{subfigure}[t]{.24\textwidth}
    \caption{}
    \centering \includegraphics[width=\textwidth]{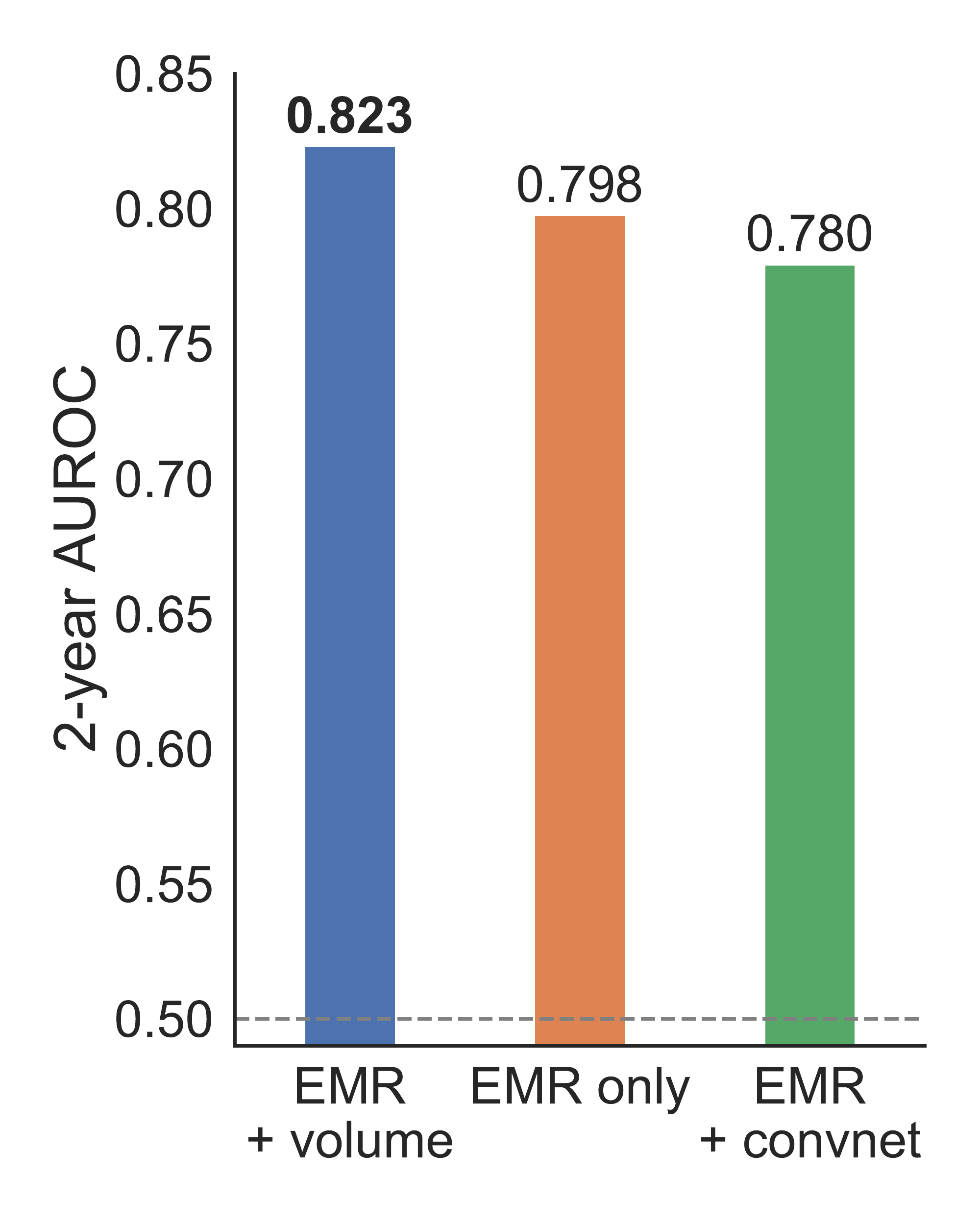}
      \label{fig:best_submission_metrics}
  \end{subfigure}
  \caption{\textbf{Winning submission.} \textbf{(a)} Overview of Deep MTLR. The
    model combines EMR features with tumour
    volume using a neural network and learns to jointly predict the probability
    of death at all intervals on the discretized time axis, allowing it to
    achieve good performance in both the binarized and lifetime risk prediction
    tasks. A predicted survival curve can be constructed for each individual to
    determine the survival probability at any timepoint. \textbf{(b)} Importance
    of combined input data for performance on the binary endpoint. Training the deep
    MTLR on EMR features only led to notably worse performance. Furthermore,
    using a deep convolutional neural network in place of tumour volume did not
    improve the 2-year AUROC.}
  \label{fig:best_submission}
\end{figure}

The winning submission (number 1) combined EMR
data with a simple image-derived measure, and used a ML model
tailored to survival prediction; a schematic overview of the submission is shown
in fig.~\ref{fig:best_submission}. The approach is based on multi-task logistic
regression (MTLR), first proposed by Yu et al~\autocite{yu_learning_2011}. In contrast
with other approaches, which focused on the binary endpoint only, MTLR is able
to exploit time-to-event information by fitting a sequence of dependent logistic
regression models to each interval on a discretized time axis, effectively
learning to predict a complete survival curve for each patient in multi-task
fashion. By making no restrictive assumptions about the form of the survival
function, the model is able to learn flexible relations between covariates and
event probability that are potentially time-varying and non-proportional. We
note that many prognostic models in clinical and radiomics literature use
proportional hazards (PH) models~\autocite{aerts_decoding_2014,
  mukherjee_shallow_2020}; however, this ignores the potential time-varying
effect of features which MTLR is able to learn. Notably, when compared to the
second-best submission (which relies on a PH model) it achieves superior
performance for lifetime risk prediction (\(C=.801\) vs \(.746\)). The added
flexibility and information-sharing capacity of multi-tasking also enables MTLR
to outperform other submissions on the binary task (\(\mathrm{AUROC}=.823\),
\(\mathrm{AP}=.505\)), even though it is not explicitly trained to maximize
predictive performance at 2 years; the predicted probabilities are also better
calibrated (Supplementary Fig.~\ref{fig:calibration_curves}). The winning
approach relies on high-level EMR features which are widely-used, easy to
interpret and show strong univariate association with survival (see
Supplementary Material). The participant incorporated non-linear interactions by
passing the features through a single-layer neural network with exponential
linear unit (ELU) activation~\autocite{fotso_deep_2018,clevert_fast_2016}, which resulted in
better performance in the development stage. The only image-derived feature
used is primary tumour volume, a known prognostic factor in HNC. Using EMR
features only, led to a decrease in performance
(\(\mathrm{AUROC}=.798\), \(\mathrm{AP}=.429\)), as did replacing tumour volume with deep image
representations learned by a 3D convnet (\(\mathrm{AUROC}=.766\),
fig.~\ref{fig:best_submission_metrics}).

\subsection*{Ensemble of all models achieves improved performance, with largest gain from a deep learning model}
\begin{figure}[H]
  \centering
  \begin{subfigure}[t]{.48\textwidth}
    \caption{}
    \centering \includegraphics[width=\textwidth]{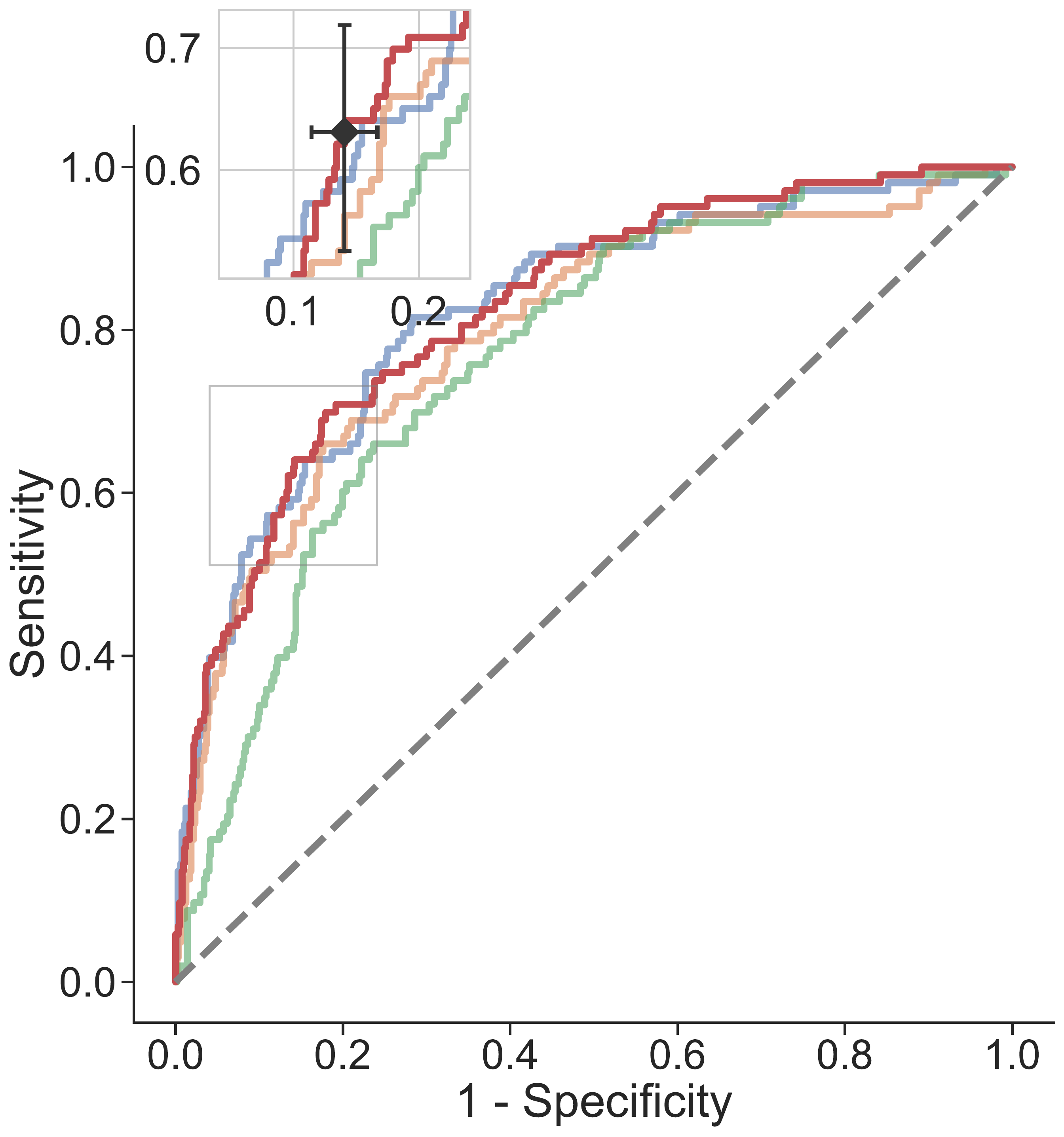}
  \end{subfigure}
  \hfill
  \begin{subfigure}[t]{.48\textwidth}
    \caption{}
    \centering \includegraphics[width=\textwidth]{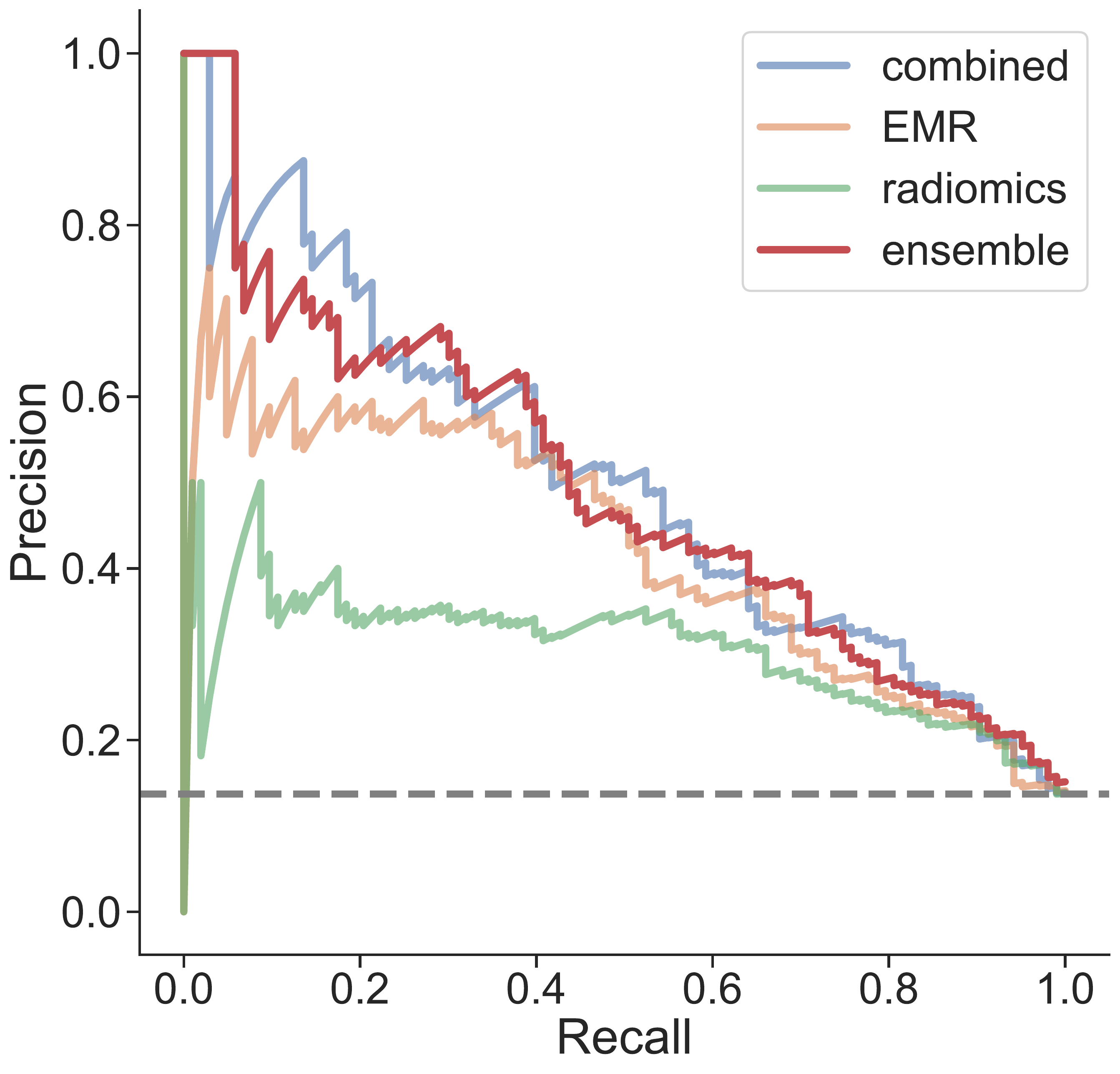}
  \end{subfigure}
  \vfill
  \begin{subfigure}[t]{.48\textwidth}
    \caption{}
    \centering \includegraphics[width=\textwidth]{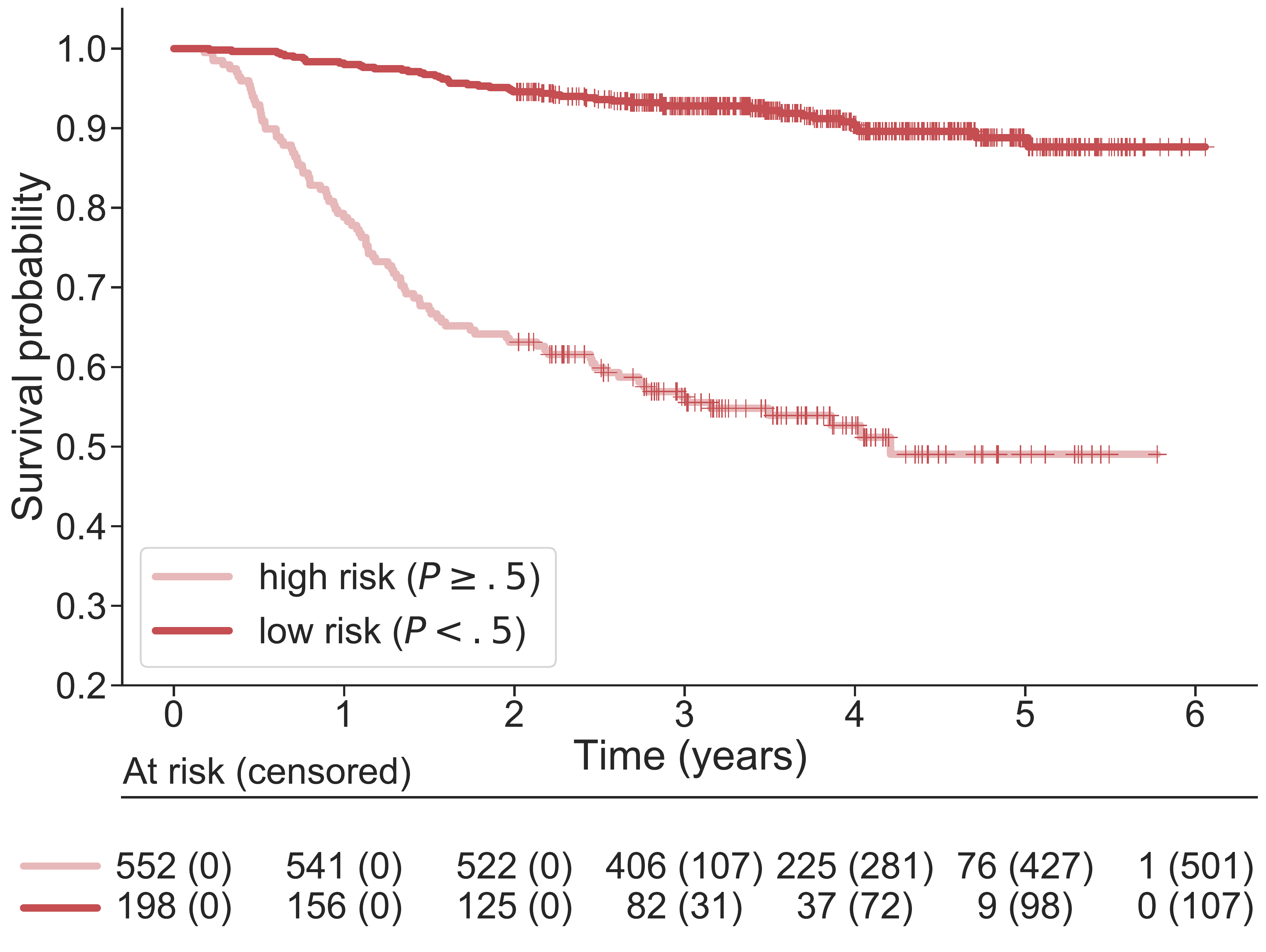}
    \vfill
    \label{fig:ensemble_km}
  \end{subfigure}
  \hfill
  \begin{subfigure}[t]{.48\textwidth}
    \caption{}
    \centering \includegraphics[width=\textwidth]{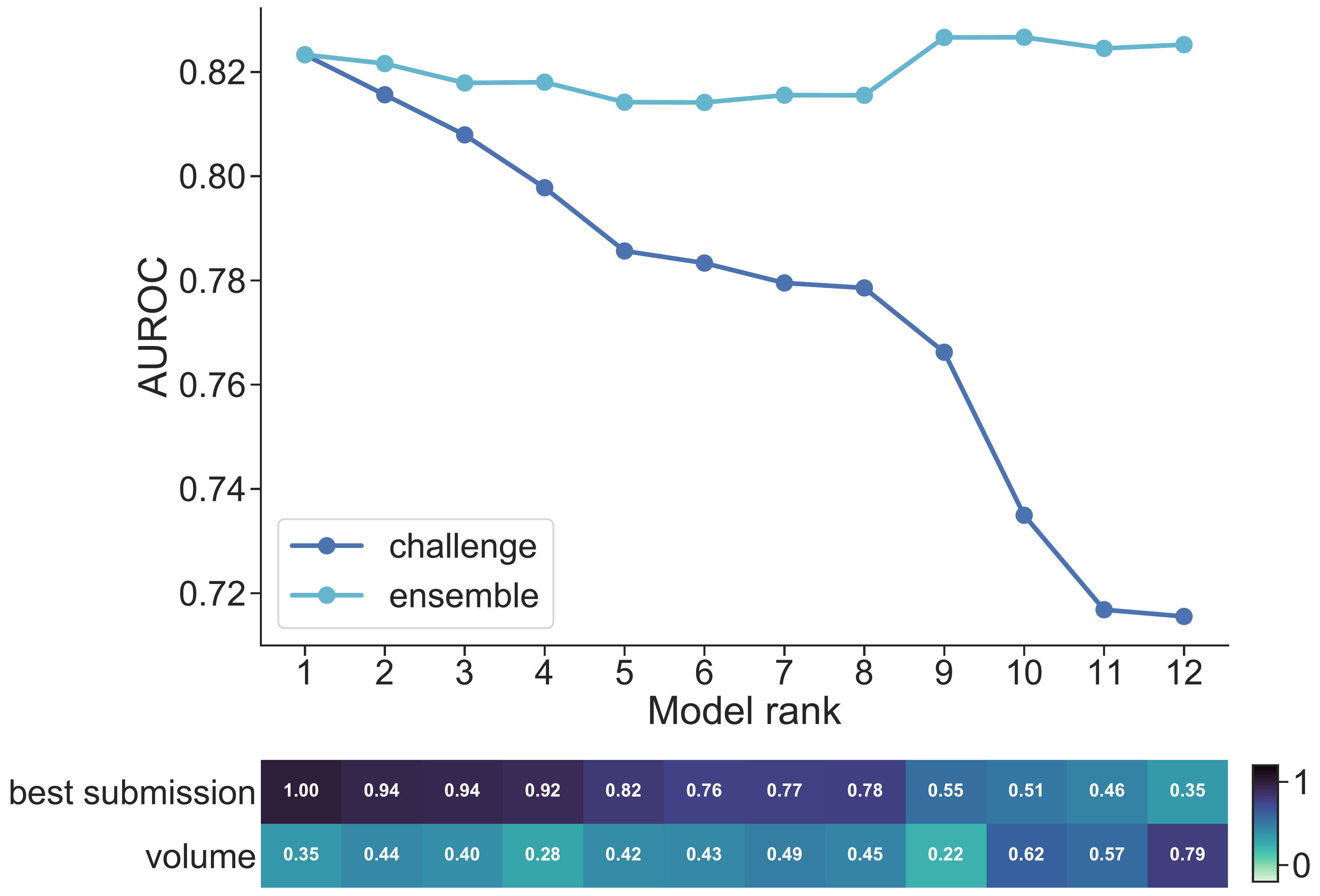}
    \label{fig:ensemble_models}
  \end{subfigure}
  \vfill
  \begin{subfigure}[t]{\textwidth}
    \caption{}
    \footnotesize
    \centering
    \resizebox{.7\columnwidth}{!}{%
      \begin{tabular}{llll}
          \toprule
          kind &                 AUROC &                    AP &               C-index \\
          \midrule
          \textbf{ensemble} &  \textbf{0.825 [0.781--0.866]} &  0.495 [0.414--0.591] &  \textbf{0.808 [0.770--0.843]} \\
          combined &  0.823 [0.777--0.866] &  \textbf{0.505 [0.420--0.602]} &  0.801 [0.757--0.842] \\
          EMR &  0.798 [0.748--0.845] &  0.429 [0.356--0.530] &  0.785 [0.740--0.827] \\
          radiomics &  0.766 [0.718--0.811] &  0.357 [0.289--0.455] &  0.748 [0.703--0.790] \\
          baseline-clinical &  0.739 [0.686--0.792] &  0.366 [0.295--0.463] &  0.696 [0.644--0.746] \\
          baseline-volume &  0.713 [0.657--0.767] &  0.317 [0.252--0.401] &  0.706 [0.653--0.756] \\
          baseline-radiomics &  0.712 [0.655--0.769] &  0.331 [0.264--0.420] &  0.729 [0.675--0.781] \\
          \bottomrule
    \end{tabular}%
    }
    \label{fig:ensemble_performance}
  \end{subfigure}
  \caption{\textbf{Performance of the ensemble approach.} \textbf{a)} ROC and
    \textbf{b)} precision-recall curves of the ensemble model and the best
    challenge submission in each category. Sensitivity and (1 - specificity) at
    the operating point chosen for risk stratification are indicated with
    bootstrap 95\% confidence intervals. \textbf{c)} Kaplan-Meier survival
    curves of the low (green) and high risk (red) subgroups identified by the
    ensemble model. \textbf{d)} Contribution of individual submissions to
    ensemble performance. Starting from the highest-ranking submissions, subset
    ensembles were created from progressively lower-ranking models. Rank=12
    corresponds to the full ensemble AUROC. The individual AUROC values are
    shown for reference. The heatmap shows the correlation of predictions of
    each individual model with best submission predictions (top) and tumour
    volume (bottom). \textbf{e)} Performance comparison between the ensemble,
    best submission from each category and baseline models.}
  \label{fig:ensemble}
\end{figure}

Ensembling, i.e., combining predictions of multiple independent models, is known
to yield superior performance to the individual predictors, as their errors tend
to cancel out while correct predictions are
reinforced~\autocite{goos_ensemble_2000}. We created an ensemble model from all
submissions by averaging the submitted predictions. In the case of missing
survival endpoint predictions, we used the 2-year survival probability as a
proxy for overall risk to compute the \(C\)-index. The ensemble model achieves
better performance in both 2-year and lifetime risk prediction
(\(\mathrm{AUROC}=.825\), \(\mathrm{AP}=.495\), \ \(C=.808\)) than any of the
individual submissions (fig.~\ref{fig:ensemble}). To investigate the risk
stratification capacity, we split the test patients into 2 groups based on the
ensemble predictions at .5 threshold. The ensemble approach was able to stratify
the patients into low and high-risk groups with significantly different survival rates
(hazard ratio 7.04, \(p<10^{-27}\), fig.~\ref{fig:ensemble_km}) and placed nearly
40\% of cancer-specific deaths in the top risk decile. Moreover, the predictions
remained significantly discriminative even when adjusted for disease site
(\(p<10^{-32}\)). It is, to our knowledge, the best published prognostic model for
overall survival in HNC using EMR and imaging data. We examined the individual
contributions to the ensemble performance by creating partial ensembles of
progressively lower-ranking submissions (fig.~\ref{fig:ensemble_models}).
Interestingly, adding the best radiomics submission (number 9) seems to provide
the greatest performance improvement. Its predictions shows low correlation with
volume and only moderate correlation with the winning model's predictions,
suggesting it might be learning prognostic signal distinct from volume and EMR
characteristics.

\section*{Discussion}
We presented the results of a machine learning challenge for HNC survival prediction. The key
strength of the proposed framework, inspired by the NCI DREAM challenges for
drug activity prediction~\autocite{nci_dream_community_community_2014}, is the
possibility of rigorous and reproducible evaluation of multiple ML approaches in
a large, multi-modal dataset~\autocite{costello_seeking_2013}. Additionally, the different participants' academic
backgrounds and computational approaches used resulted in a diverse collection
of models and made it possible to build a 'wisdom of the crowds' ensemble model
with improved performance.

The best individual approach achieved strong performance on both 2-year and
lifetime risk prediction using a multi-task survival modelling framework. This
demonstrates the benefit of using a flexible approach designed specifically for the task of survival prediction. Additionally, because the approach relies on widely
used and easy-to-interpret features (e.g. tumour stage, volume), it is
attractive from a clinical standpoint as a risk stratification and monitoring
tool. Our best ensemble model combined the strengths of all submissions to
establish, to our knowledge, a new state-of-the-art result for HNC survival
prediction. The model predictions are highly significant, even when adjusted for
disease site, demonstrating the potential of learning from large cross-sectional
datasets, as opposed to highly curated patient subsets (which has been the
dominant paradigm thus far).

The utility of radiomics in HNC survival prediction has been investigated in recent
studies~\autocite{vallieres_radiomics_2017, diamant_deep_2019, ger_radiomics_2019}.
We have identified several strong radiomics predictors; however, the best
performing individual submission used EMR features,
with primary tumour volume as the only image-derived feature. Our conclusions
match those of Ger et al.~\autocite{ger_radiomics_2019}, who did not find significant
improvement in prognostic performance of handcrafted CT and PET imaging features
in HNC compared to volume alone and of Valli\`{e}res et al.~\autocite{vallieres_radiomics_2017}, whose
best performing model for overall survival also combined EMR features and volume. We
further showed that although deep learning-based imaging models generally
outperformed approaches based on handcrafted features, none proved superior to
the combined EMR-volume model, even when combined with
EMR data. Deep learning methods achieve excellent
performance in many image processing tasks~\autocite{kolesnikov_big_2020}, however,
current approaches require substantial amounts of training data.

While our dataset is the largest publicly-available HCN imaging collections, it
is still relatively small compared to natural image datasets used in ML
research, which often contain millions of
samples~\autocite{sun_revisiting_2017}. Although such large sample sizes might
be unachievable in this particular setting, better data collection and sharing
practices can help build more useful databases (the UK
Biobank~\autocite{sudlow_uk_2015} or the Cancer Genome
Atlas~\autocite{the_cancer_genome_atlas_research_network_cancer_2013} are
excellent examples). This is especially important in diseases with low event
rates, where a substantial number of patients might be needed to capture the
variation in phenotype and outcomes. The inferior performance of radiomic models
can also be attributed to suboptimal imaging data. While the possibility to
easily extract retrospective patient cohorts makes routine clinical images
attractive for radiomics research, they are often acquired for purposes entirely
orthogonal to new biomarker discovery. CT images in particular might not accurately reflect
the biological tumour characteristics due to insufficient resolution,
sensitivity to acquisition parameters and noise~\autocite{fave_preliminary_2015,
  hatt_characterization_2017}, as well as the source of image contrast, which is essentially electron density of the tissue which demonstrates little texture at current image scales. This highlights the broader need of greater
collaboration between ML researchers, clinicians and physicists, also in data
selection and experiment design --- with reciprocal feedback~\autocite{mateen_improving_2020, kazmierska_multisource_2020}.

Our study has several potential limitations. Participation was restricted to
one institution, which limited the number of submissions we received.
Additionally, the hand-engineered radiomics submissions relied on one radiomics
toolkit (PyRadiomics) and other widely-used toolkits make use of potentially
different feature sets and definitions; however, thanks to recent efforts in
image biomarker standardization, the features have been shown to be largely
consistent between the major implementations~\autocite{zwanenburg_image_2020}.
Additionally, our dataset was collected within one hospital only. Assessing
generalizability of the best performing models to other institutions and patient
populations is important for future clinical implementation. This is
particularly relevant for radiomics models as domain shift due to differences in
scanning equipment and protocols could negatively affect generalization and we are currently working on validating
the best performing models using multi-institutional data. It is
likely that more sophisticated ensembling methods (e.g. Bayesian model
averaging~\autocite{goos_ensemble_2000} or stacking~\autocite{wolpert_stacked_1992})
could achieve even better performance by weighing the models according to
their strengths. We leave this exploration for future work.

In the future, we would like to open the challenge to a larger number of
participants, which would further enhance the diversity of approaches and help
us validate our conclusions. We are also working on collecting additional
outcome information, including recurrence, distant metastasis and treatment
toxicity, which would provide a richer set of prediction targets and might be
more relevant from a clinical standpoint. The importance of ML and AI as tools
of precision medicine will continue to grow. However, it is only through transparent and reproducible research that integrates diverse knowledge that we can begin to realize the full potential of these
methods and permit integration into clinical practice.

\section*{Acknowledgements}
We would like to thank the Princess Margaret Head and Neck Cancer group for the support with data collection and curation. We would also like to acknowledge Zhibin Lu and the HPC4Health team for the technical support. MK is supported by the Strategic Training in Transdisciplinary Radiation Science for the 21st Century Program (STARS21) scholarship.

\section*{Methods}

\subsection*{Dataset}
We collected a retrospective dataset of 2552 HNC patients treated with
radiotherapy or chemoradiation at PM Cancer Centre between 2005 and 2017 (Supplementary table \ref{tab:data_overview}), which we split into training and test subsets by date of diagnosis (2005-2015 and 2016-2018 for training and
independent test set, respectively). The
study was approved by the institutional Research Ethics Board (REB \#17-5871). The
inclusion criteria were: 1) availability of planning CT image and target
contours; 2) at least 2 years follow-up (or death before that time); and 3) no
distant metastases at diagnosis and no prior surgery. Primary gross tumour
volumes (GTV) were delineated by radiation oncologists as part of
routine treatment planning. For each patient we exported the CT image
and primary GTV binary mask in NRRD format. We also extracted the follow-up
information (current as of April 2020). The dataset was split into training
(\(n=1802\)) and test (\(n=750\)) subsets according to the date of diagnosis (Supplementary fig. \ref{fig:data_curation_flowchart}).

\subsection*{Baseline models}
To provide baselines for comparison and a reference point for the participants,
we created three benchmark models using: 1) standard prognostic factors used in
the clinic (age, sex, T/N stage and HPV status) (\texttt{baseline-clinical}); 2)
primary tumour volume only (\texttt{baseline-volume}); 3) handcrafted imaging
features (\texttt{baseline-radiomics}). All categorical variables were one-hot encoded and missing data was handled by creating additional category representing missing value (e.g. 'Not tested' for HPV status). For the \texttt{baseline-radiomics}
model, we extracted all available first order, shape and textural features from
the original image and all available filters (1316 features in total) using the
PyRadiomics package (version 2.2.0)~\autocite{van_griethuysen_computational_2017} and performed
feature selection using maximum relevance-minimum redundancy (MRMR) method~\autocite{hanchuan_peng_feature_2005}. The number of selected features and model
hyperparameters (\(l_2\) regularization strength) were tuned using grid search
with 5-fold cross validation. All models, were built using logistic regression
for the binary endpoint and a proportional hazards model for the survival
endpoint.

\subsection*{Participant resources}
All participants had access to a shared Github repository containing the code used for data processing, as well as implementations of the baseline models and example modelling pipelines, facilitating rapid development process. A dedicated Slack workspace was used for announcements, communication with organizers and between participants and to share useful learning resources. We hosted the dataset on an institutional high-performance
computing (HPC) cluster with multicore CPUs and general-purpose graphics processing units (GPUs) which were available for model training.

\subsection*{Tasks and performance metrics}
The main objective of the challenge was to predict binarized 2-year overall
survival (OS), with the supplementary task of predicting lifetime risk of death
and full survival curves (in 1-month intervals from 0 to 23 months). To evaluate
and compare model performance on the 2-year binarized survival prediction task,
we used area under the ROC curve (AUROC), which is a ranking metric computed
over all possible decision thresholds. We additionally computed the area under
precision-recall curve, also referred to as average precision (AP), using the
formula:
\[
\mathrm{AP} = \sum_n (R_n - R_{n-1})P_n,
\]
where \(R_n\) and \(p_n\) are the precision and recall at a given threshold,
respectively. While AUROC is insensitive to class balance, AP considers the
positive class only, which can reveal pathologies under high class imbalance~\autocite{lever_classification_2016}. Additionally, both metrics consider all
possible operating points, which removes the need to choose a particular
decision threshold (which can vary depending on the downstream clinical task).
Since the dataset did not include patients with follow-up time less than 2
years, we did not correct the binary metrics for censoring bias.

For the lifetime risk prediction task, we used concordance index,
defined as:
\[
C = \frac{\sum_{i \text{ uncensored}}\sum_{t_j > t_i}\mathbf{1}\{r_i > r_j\} +
  \frac{1}{2}\mathbf{1}\{r_i = r_j\}}{\sum_{i \text{ uncensored}}\mathbf{1}\{t_j > t_i\}},
\]
where \(t_i\) is the time until death or censoring for patient \(i\), \(r_i\) is
the predicted risk score for patient \(i\) and \(\mathbf{1}\{\}\) is the indicator function. The agreement between the performance
measures was good (Pearson \(r=.88\) between AUROC and AP, \(r=.82\)
between AUROC and \(C\)-index). We compared the AUROC achieved by the best submission to the other submissions using one-sided \(t\)-test and corrected for multiple comparisons by controlling the false discovery rate (FDR) at .05 level.

\subsection*{Research reproducibility}
The code used to prepare the data, train the baseline models, evaluate the challenge submissions and analyze the results is available on Github at \url{https://github.com/bhklab/uhn-radcure-challenge}. We also share the model code for all the challenge submissions with the participants' permission in the same repository. Furthermore, we are planning to make the complete dataset, including anonymized images, contours and EMR data available on the Cancer Imaging Archive.

\printbibliography

\pagebreak
\appendix
\renewcommand{\thefigure}{S\arabic{figure}}
\renewcommand{\thetable}{S\arabic{figure}}
\setcounter{figure}{0}
\section*{Supplementary material}
\subsection*{Data curation and preprocessing}
\begin{figure}[H]
  \centering
  \includegraphics[width=.9\textwidth]{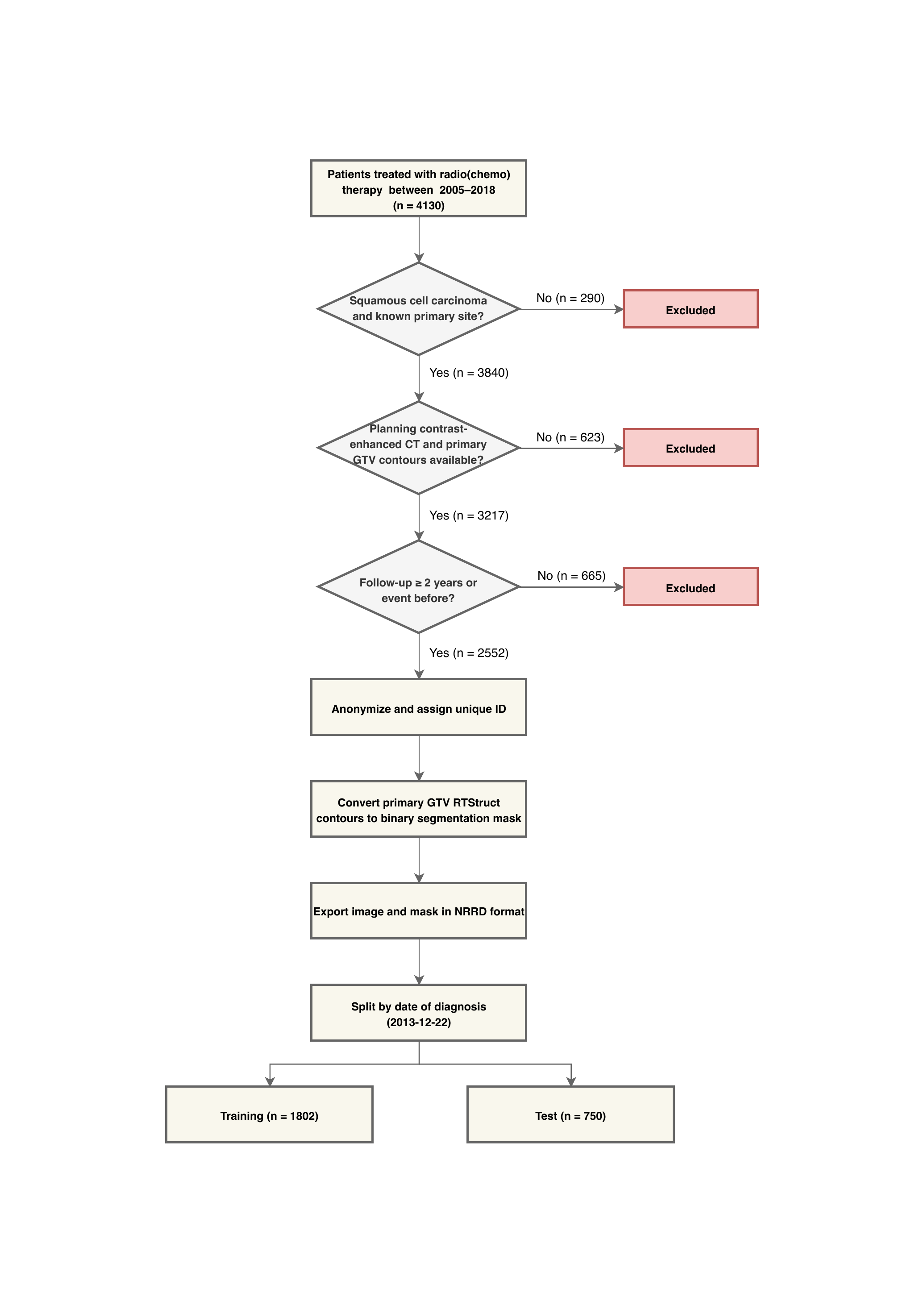}
  \caption{Patient selection and data curation process.}
  \label{fig:data_curation_flowchart}
\end{figure}

\subsection*{Patient characteristics}
\newcommand{\rowgroup}[1]{\hspace{-1em}#1}
\begin{table}[H]
\centering
\begin{tabular}{>{\quad}lrr}
\toprule
& \textbf{Training} & \textbf{Test} \\
\midrule
\rowgroup{\# of patients} & 1802 & 750\\

\rowgroup{Outcome}\\
Alive/Censored & 1065 (59\%) & 609 (81\%)\\
Dead & 737  (41\%) & 141 (19\%)\\
\quad Dead before 2 years & 323 (18\%) & 103 (14\%)\\

\midrule
\rowgroup{Sex}\\
Male    & 1424 (79\%)  & 615 (82\%)\\
Female  & 378 (21\%)  & 135 (18\%)\\

\midrule
\rowgroup{Disease Site}\\
Oropharynx          & 777 (43\%) & 399 (53\%)\\
Larynx              & 555 (31\%) & 172 (23\%)\\
Nasopharynx         & 220 (12\%) & 101 (13\%)\\
Hypopharynx         & 115 (6.4\%)& 28 (3.7\%)\\
Lip / Oral Cavity   & 72 (4.0\%) & 10 (1.3\%)\\
Nasal Cavity        & 31 (1.7\%) & 20 (2.7\%)\\
Paranasal Sinus     & 16 (0.9\%) & 10 (1.3\%)\\
Esophagus           & 14 (0.8\%) & 8 (1.1\%)\\
Salivary Glands     & 2 (0.1\%)  & 2 (0.3\%)\\

\midrule
\rowgroup{T stage}\\
1/2   & 919 (51\%)  & 405  (54\%)\\
3/4  & 855 (47\%)  & 336  (45\%)\\
Not available   & 28 (2\%)   & 9 (1\%)\\

\midrule
\rowgroup{N stage}\\
0   & 688 (38\%)  & 226  (30\%)\\
1  & 170 (9\%)  & 90  (12\%)\\
2 & 835 (46\%)  & 395 (53\%)\\
3  & 108 (6\%)  & 39 (5\%)\\
Not available   & 1 (\textless1\%)   & 0 (0\%)\\

\midrule
\rowgroup{AJCC stage}\\
I/II   & 455 (25\%)  & 158  (21\%)\\
III/IV & 1318 (73\%)  & 581 (77\%)\\
Not available   & 29 (2\%)   & 11 (1\%)\\

\midrule
\rowgroup{HPV status}\\
Positive   & 513 (28\%)  & 327  (44\%)\\
Negative  & 269 (15\%)  & 168  (22\%)\\
Not tested   & 1020 (57\%)   & 255 (34\%)\\

\midrule
\rowgroup{Chemotherapy}\\
Yes   & 725 (40\%)  & 361  (48\%)\\
No  & 1077 (60\%)  & 389  (52\%)\\

\midrule
\rowgroup{ECOG performance status}\\
0   & 1120 (62\%)  & 436  (58\%)\\
1 & 489 (27\%)  & 290 (39\%)\\
2 & 145 (8\%)  & 20 (3\%)\\
\textgreater2 & 34 (2\%)  & 4 (1\%)\\
Not available   & 14 (1\%)   & 0 (0\%)\\
\bottomrule
\end{tabular}
\caption{Patient characteristics in the training and test sets.}
\label{tab:data_overview}
\end{table}

\subsection*{Imaging protocols}
\begin{table}[H]
\centering
\begin{tabular}{ll}
\toprule
Parameter & Median [range]\\
\midrule
Slice thickness & 2 [2--2.5]\\
kVp & 120 [120--120]\\
Tube current [mAs] & 300 [121--540]\\
Pixel spacing [mm] & (.976, .976) [(.702, .702)--(1.17, 1.17)]\\
\bottomrule
\end{tabular}
\caption{CT imaging parameters used in the study. Images were acquired using either of: Toshiba Aquilion One (\(n=1653\)), GE Discovery ST (\(n=643\)), GE LightSpeed Plus (\(n=130\)), Philips Brilliance Big Bore (\(n=96\)) or GE Discovery 610 (\(n=30\)).}
\end{table}

\begin{refsection}
\subsection*{Detailed descriptions of submissions}
\subsubsection*{Submission 1}
Multi-task logistic regression (MTLR) uses a sequence of dependent regressors to
predict the probability of event occuring at multiple discrete timepoints in a
multi-task fashion~\autocite{yu_learning_2011}. The model is trained by minimizing the MTLR
log-likelihood:

\begin{align*}
  L(\boldsymbol{\Theta}, D) &=
                              \sum_{j=1}^{N-N_c-1}\sum_{k=1}^{K-1}(\boldsymbol{\theta}_k^T \mathbf{\phi(x)}^{(j)} +
                              b_k)y_k^{(j)} &\text{(Uncensored)}\\
                            &+ \sum_{j=N-N_c}^{N}\log(\sum_{i=1}^{K-1}\mathbf{1}\{t_i \ge T_c^{(j)}\}\exp(\sum_{k=i}^{K-1}((\boldsymbol{\theta}_k^T\mathbf{\phi(x)}^{(j)} + b_k)y_k^{(j)}))) &\text{(Censored)}\\
                            &- \sum_{j=1}^{N}\log(\sum_{i=1}^{K}
                              \exp(\sum_{k=i}^{K-1}\boldsymbol{\theta}_k^T \mathbf{\phi(x)}^{(j)} + b_k)), &\text{(Normalizing constant)}\\
                            &+ \frac{C_1}{2}\sum_{k=1}^{K-1}\Vert\boldsymbol{\theta}_k\Vert_2^2 &\text{(Regularizer)},
\end{align*}
where \((\boldsymbol{\theta}_k,\ b_k)\) are trainable parameters associated with
the \(k\)th timepoint, \(D = \{T^{(j)}, \delta^{(j)},
\mathbf{x}^{(j)}\}_{j=1}^{N}\) is a dataset with \(N\) patients, \(N_c\) of whom
are censored and \(y_k\) is the binary event indicator for timepoint \(k\).
We define \(\phi(x)\) to be a multi-layer perceptron with exponential linear
unit (ELU) activations~\autocite{fotso_deep_2018,clevert_fast_2016}.

The model takes a vector of EMR features and volume as input and predicts the
probability of event occurring within each of the discrete time intervals. An
individual survival curve can be computed for each patient from the predictions,
from which we read out the probability of 2-year survival, as well as the
lifetime risk score. Tumour volume was computed using the mesh algorithm
implemented in PyRadiomics version 2.2.0. We used one-hot encoding for
categorical EMR features and encoded any missing values as separate dummy
category. The inputs were normalized to zero mean and unit variance using
statistics computed on the training set. We implemented the MTLR algorithm using
using the PyTorch framework version 1.5.0 and trained it for 100 epochs using
the Adam optimizer with default momentum
parameters~\autocite{kingma_adam:_2014}. The number of MTLR time bins was set as
\(\sqrt{N_{\mathrm{uncensored}}}\), where \(N_{\mathrm{uncensored}}\) is the number of uncensored patients in the training set and the bin edges were set to
quantiles of training survival time distribution. We tuned other hyperparameters
by maximizing 5-fold cross validation AUROC on the training set using 60
iterations of random search. The final hyperparameter settings are shown in the
table below.
\begin{table}[H]
  \centering
  \begin{tabular}{ll}
    \toprule
    Hyperparameter & Value\\
    \midrule
    Batch size & 512\\
    Dropout & .24\\
    Hidden layer sizes & (128)\\
    \(C_1\) & 10\\
    Learning rate & .006\\
    Weight decay & \(6 \times 10^{-5}\)\\
    \bottomrule
  \end{tabular}
  \caption{Submission 1 hyperparameters.}
\end{table}

\subsubsection*{Submission 2}
We applied the fuzzy prediction framework used by submission 4 to EMR variables
only (no engineered radiomics except tumour volume used as the fuzzy variable).

\subsubsection*{Submission 3}
We attempted to identify prognostic signal in radiomic features beyond tumour
volume using a fuzzy learning approach~\autocite{haibe-kains_fuzzy_2010}.
Briefly, we divided the training dataset into 2 groups based on the tumour
volume (greater/less than median) and trained a binary logistic regression model
to predict the probability of falling in the large volume group from the input
features. We then built a separate logistic regression (for the binary task) and
Cox proportional hazard (for the lifetime survival task) within each of the
subgroups. The final predictions were computed as a linear combination of the
subgroup model predictions weighted by the predicted probability of falling
within the subgroup. We used the provided EMR features as inputs together with
hand-engineered imaging features. All categorical features were converted to
indicator variables (with a separate category for missing values) and continuous
features were normalized to zero mean and unit variance. To reduce the
dimensionality of feature space and prevent the inclusion of potentially noisy
variables, we used a curated set of radiomic features which were previously
found to be prognostic in HNC: \texttt{GLSZM-SizeZoneNonUniformity},
\texttt{GLSZM-ZoneVariance},
\texttt{GLRLM-LongRunHighGrayLevelEmphasis}\autocite{vallieres_radiomics_2017,diamant_deep_2019}.
We extracted the features using PyRadiomics 2.2.0 with fixed quantization bin
width equal to 25 and after resampling the images to isotropic 1mm voxel
spacing. Standard logistic regression (as implemented in Scikit-learn
package~\autocite{pedregosa_scikit-learn:_2011}, version 0.22.1) with LBFGS
solver and inverse frequency weighted loss function was used for binarized
output, Cox modeling (Lifelines package, version 0.24.5) with partial hazard
fitting and step size of 0.5 was used for survival prediction.

\subsubsection*{Submission 4}
The same algorithm as submission 1 but with EMR data only and different network
architecture.
\begin{table}[H]
  \centering
  \begin{tabular}{ll}
    \toprule
    Hyperparameter & Value\\
    \midrule
    Batch size & 1024\\
    Dropout & .14\\
    Hidden layer sizes & (32, 32, 32)\\
    \(C_1\) & 10\\
    Learning rate & .006\\
    Weight decay & \(1.3 \times 10^{-6}\)\\
    \bottomrule
  \end{tabular}
  \caption{Submission 3 hyperparameters.}
\end{table}

\subsubsection*{Submission 5}
Our approach uses a 3D convnet with EMR features concatenated before
fully-connected layers. The convnet uses \texttt{conv-batch norm-leaky ReLU}
block structure with negative activation slope equal to .1. The network takes a
\(50\mathrm{mm}\times 50\mathrm{mm}\times 50\mathrm{mm}\) image patch centred on the GTV mask
centroid as input and outputs the predicted probability of death before 2 years.
The images were resampled to isotropic 1mm spacing, intensity clipped to [-500
HU, 1000 HU] range and normalized by subtracting the training set mean and
dividing by training standard deviation. EMR features were normalized
analogously and categorical features were additionally one-hot encoded,
replacing any missing values with a special 'missing' category. We applied
random data augmentation, including random in-plane rotations between \((-\pi/6,
\pi/6)\) radians, random flipping along the \(x\) and \(z\) axes and additive Gaussian
noise with standard deviation .05 (after normalization). We implemented the
model using PyTorch 1.5.0 and PyTorch Lightning 0.7.6 and trained it for 500
epochs using the Adam algorithm with batch size 10 and learning rate \(10^{-3}\)
decayed by a factor of 10 after 60, 160 and 360 epochs. To prevent overfitting,
we used dropout with probability .4 and weight decay regularization with
coefficient \(10^{-4}\). We used binary cross entropy loss weighted by the
inverse frequency of positive label to reduce the impact of class imbalance. The
hyperparameters were selected based on performance on a 10\% validation set held
out from the training set.

\begin{table}[H]
  \centering
  \begin{tabular}{lcc}
    \toprule
    Operation & Output channels & Kernel size\\
    \midrule
    Convolution & 64 & \(5^3\)\\
    Convolution & 128 & \(3^3\)\\
    Max pooling (stride=2) & - & \(2^3\)\\
    Convolution & 256 & \(3^3\)\\
    Convolution & 512 & \(3^3\)\\
    Max pooling (stride=2) & - & \(2^3\)\\
    Global average pooling & - & -\\
    Fully-connected & 512 & -\\
    Concatenate EMR features & - & -\\
    Fully-connected & 512 & -\\
    Dropout (p=.4) & - & - \\
    Fully-connected & 1 & -\\
    \bottomrule
  \end{tabular}
  \caption{Convnet architecture. Each convolution operation corresponds to the
    block described above.}
\end{table}

\subsubsection*{Submission 6}
We used a 2D convnet analogously to submission 10 and combined the learned image
features with EMR variables. The EMR features were one-hot encoded and
normalized to zero mean and unit variance before being passed through three
fully-connected layers with 8 hidden units and concatenated with the convnet
output before the final classification layer.

\subsubsection*{Submission 7}
To learn prognostic image representations, we used a 3D dense convolutional
network (DenseNet) with multitask learning prediction head. The network takes a
cropped \(60\mathrm{mm}\times 60\mathrm{mm}\times 60\mathrm{mm}\) image patch,
centred on the GTV centroid and outputs the probability of death at multiple
discrete time intervals. We used convolutional block structure previously
validated in retinal tomography scans~\autocite{de_fauw_clinically_2018}. Each
convolutional block consists of multiple layers of within-slice (\(1 \times 3 \times 3\))
and across-slice (\(3 \times 1 \times 1\)) convolutions, followed by batch
normalization~\autocite{ioffe_batch_2015} and ReLU nonlinearities. The EMR
features were concatenated with the convnet output before the final prediction
layer. The convnet architecture is shown in table \ref{tab:densenet} below.

\begin{table}[H]
  \centering
  \begin{tabular}{lc}
    \toprule
    Block & Operations \\
    \midrule
    Convolution & \(\begin{bmatrix} \mathrm{conv}\ 1\times 3 \times 3 \\ \mathrm{conv}\ 1\times 3 \times 3 \end{bmatrix}\)\\
    Pooling & max pool \(2\times 2\times 2\), stride 2\\
    \cmidrule{1-2}
    Dense block 1 & \(\begin{bmatrix}\mathrm{conv} 1\times 3\times 3 \\ \mathrm{conv} 1\times 3\times 3 \\ \mathrm{conv} 3\times 1\times 1 \end{bmatrix} \times 2\)\\
    \cmidrule{1-2}
    \multirow{2}{*}{Transition 1} & conv \(1\times 1\times 1\) \\ & max pool \(2\times 2\times 2\), stride 2\\
    \cmidrule{1-2}
    Dense block 2 & \(\begin{bmatrix}\mathrm{conv} 1\times 3\times 3 \\ \mathrm{conv} 1\times 3\times 3 \\ \mathrm{conv} 3\times 1\times 1 \end{bmatrix} \times 2\)\\
    \cmidrule{1-2}
    \multirow{2}{*}{Transition 2} & conv \(1\times 1\times 1\) \\ & max pool \(2\times 2\times 2\), stride 2\\
    \cmidrule{1-2}
    Dense block 3 & \(\begin{bmatrix}\mathrm{conv} 1\times 3\times 3 \\ \mathrm{conv} 1\times 3\times 3 \\ \mathrm{conv} 3\times 1\times 1 \end{bmatrix} \times 2\)\\
    \cmidrule{1-2}
    \multirow{2}{*}{Transition 3} & conv \(1\times 1\times 1\) \\ & max pool \(2\times 2\times 2\), stride 2\\
    \cmidrule{1-2}
    Dense block 4 & \(\begin{bmatrix}\mathrm{conv} 1\times 3\times 3 \\ \mathrm{conv} 1\times 3\times 3 \\ \mathrm{conv} 3\times 1\times 1 \end{bmatrix} \times 3\)\\
    \cmidrule{1-2}
    Global pool & adaptive average pool\\
    EMR features  & concat(input, EMR features)\\
    Output & MTLR(time bins=40)\\
    \bottomrule
  \end{tabular}
  \caption{3D Dense Net architecture. The convolution kernel sizes are given as
    (depth, width, height). Each conv operation above corresponds to the
    sequence batch norm-ReLU-conv, except for the first 2 convolutions
    where the order is reversed. Additionally, we applied dropout after each
    transition layer. Note that the number of channels in each layer is
    determined by the first convolution output channels (here 32) and the growth
    rate (tunable hyperparameter).}
    \label{tab:densenet}
\end{table}

The input image patches were resampled to \(3\mathrm{mm}\times
1\mathrm{mm}\times 1\mathrm{mm}\) voxel spacing and clipped to range [-500,
1000] HU. For EMR features, we one-hot encoded categorical inputs features, with
any missing values represented as separate category. Both images and EMR
features were normalized to zero mean and unit variance using statistics
computed on the training set. 10\% of training patients were set aside as a
validation set for hyperparameter tuning. We applied random data augmentation to
input image patches (table \ref{tab:augmentation}). The augmentation operations
were implemented in SimpleITK version 1.2.4 and fused into a single transform to
minimize interpolation artifacts.

\begin{table}[H]
  \centering
  \begin{tabular}{ll}
    \toprule
    Parameter & Value or range \\
    \midrule
    In-plane rotation & [-10\textdegree--10\textdegree]\\
    Flip along z-axis & [True, False]\\
    In-plane shear & [-.005, .005]\\
    In-plane scaling & [.8, 1.2]\\
    In-plane translation & [-10mm, 10mm]\\
    In-plane elastic deformation & \(\text{grid size} = (2, 2)\), \(
    \alpha=5\)\\ 
    Gaussian noise & \(\mu=0\), \(\sigma=10\)\\
    \bottomrule
  \end{tabular}
  \caption{Data augmentation operations used during training. Parameter values
    were drawn randomly for each input from the ranges shown above.}
  \label{tab:augmentation}
\end{table}

We implemented our approach using PyTorch version 1.5.0 and PyTorch Lightning
version 0.7.6. The model was trained by minimizing the MTLR negative log
likelihood using the Adam optimizer with default momentum parameters for a
maximum of 200 epochs. Training was stopped early if the loss on the validation
set did not improve by at least .0005 for 20 epochs. The initial learning rate
was decayed by a factor of .5 after 60, 100, 140 and 180 epochs. To mitigate the
high class imbalance present in the dataset, we oversampled the minority class
by using a balanced minibatch sampler, which helped to stabilize training.
Hyperparameters were selected by maximizing the validation set performance using
60 iterations of random search. The final hyperparameter configuration is shown
below.

\begin{table}[H]
  \centering
  \begin{tabular}{ll}
    \toprule
    Hyperparameter & Value\\
    \midrule
    Batch size & 8\\
    Dense block layers & (2, 2, 2, 3)\\
    DenseNet growth rate & 24\\
    Dropout & .38\\
    Initial num. channels & 32\\
    MTLR regularization & 10\\
    Learning rate & .0002\\
    Weight decay & \(9.4 \times 10^{-4}\)\\
    \bottomrule
  \end{tabular}
  \caption{Submission 7 hyperparameters.}
\end{table}

\subsubsection*{Submission 8}
We trained a three-layer neural network with scaled exponential linear unit
(SELU) activation and alpha-dropout~\autocite{klambauer_self-normalizing_2017}. The inputs were EMR features after
one-hot encoding (for categorical features) and normalization (for continuous
features) and the output was the predicted 2-year survival probability. To
address the issue of class imbalance, we used biased sampling to adjust the
frequency of positive training samples. We selected the hyperparameters manually
based on cross-validation performance.

\subsubsection*{Submission 9}
We used a 3D dense convolutional network (DenseNet) with similar block structure
and architecture as submission 7. The main differences were the lack of EMR
inputs and a second context network, taking a downsampled image patch with the
same location as the base network but \(2\times\) lower resolution, providing a
zoomed-out view of the tumour surroundings. The feature maps from both streams
were concatenated along the channel dimension before global pooling and passed
through additional \(1\times 1\times 1\) convolution to maintain equal number of
channels. The final hyperparameter configuration is shown below.

\begin{table}[H]
  \centering
  \begin{tabular}{ll}
    \toprule
    Hyperparameter & Value\\
    \midrule
    Batch size & 16\\
    Dense block layers & (2, 2, 2, 3)\\
    DenseNet growth rate & 24\\
    Dropout & .03\\
    Initial num. channels & 32\\
    MTLR regularization & 10\\
    Learning rate & .00027\\
    Weight decay & \(1.7 \times 10^{-4}\)\\
    \bottomrule
  \end{tabular}
  \caption{Submission 9 hyperparameters.}
\end{table}

\subsubsection*{Submission 10}
The architecture used was a VGGNet~\autocite{simonyan_very_2015} with batch
normalization and sigmoid output for binary classification. The inputs were
formed by extracting the largest 2D GTV slice and concatenating with the binary
mask along the channel axis. The images were cropped to \(96 \times 96\) pixel
window centred on the mask centroid, intensity clipped to [-1000, 400] HU range
and normalized to zero mean and unit variance. We applied data augmentation
including additive Gaussian noise (image channel only) with standard deviation
of 12 HU, random translations between \(\pm 20\) pixels in each direction and
random scaling between .85 and 1.25. The model was implemented in PyTorch and
trained with minority class oversampling to mitigate class imbalance.

\subsubsection*{Submission 11}
We used similar training setup and implementation as submission 5 but relying on
images only (without EMR features) and a different convnet architecture (see below).

\begin{table}[H]
  \centering
  \begin{tabular}{rcc}
    \toprule
    Operation & Output channels & Kernel size\\
    \midrule
    Convolution & 64 & \(3^3\)\\
    Convolution & 128 & \(3^3\)\\
    Convolution & 128 & \(3^3\)\\
    Max pooling (stride=2) & - & \(2^3\)\\
    Convolution & 256 & \(3^3\)\\
    Convolution & 256 & \(3^3\)\\
    Convolution & 512 & \(3^3\)\\
    Max pooling (stride=2) & - & \(2^3\)\\
    Convolution & 512 & \(3^3\)\\
    Convolution & 1024 & \(3^3\)\\
    Convolution & 1024 & \(3^3\)\\
    Max pooling (stride=2) & - & \(2^3\)\\
    Global average pooling & - &  -\\
    Fully-connected & 1024 & -\\
    Dropout (p=.4) & - & -\\
    Fully-connected & 1 & -\\
    \bottomrule
  \end{tabular}
  \caption{Convnet architecture.}
\end{table}

\subsubsection*{Submission 12}
We applied the fuzzy training framework used in submissions 2 and 3 to the
curated set of radiomic features only.

\subsection*{Additional results}
\begin{figure}[H]
  \centering
  \includegraphics[width=\textwidth]{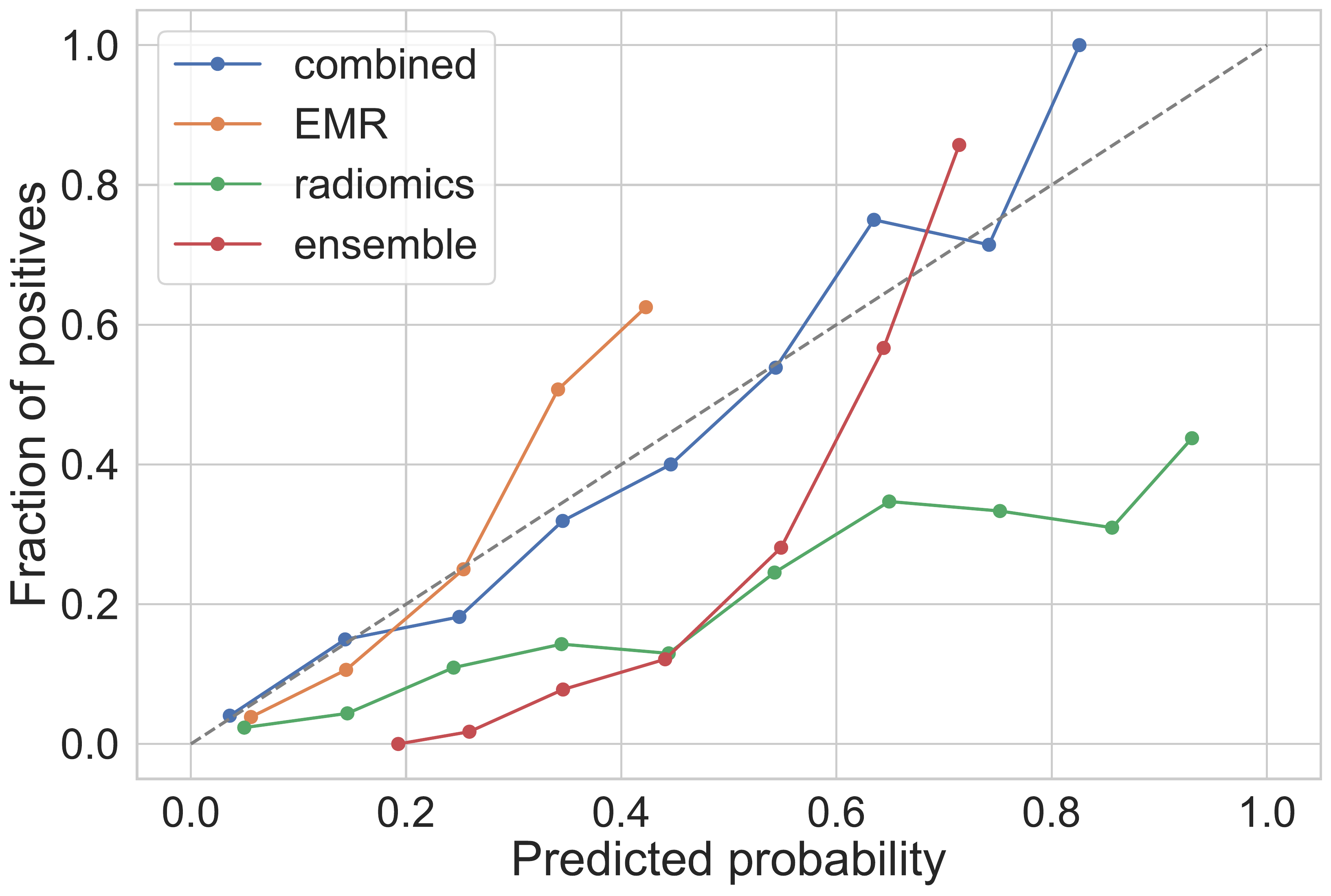}
  \caption{Calibration of predicted 2-year event probabilities for the best
    performing model in each category and the ensemble of all models.}
  \label{fig:calibration_curves}
\end{figure}

\printbibliography
\end{refsection}

\end{document}